\documentclass[10pt,twocolumn,letterpaper]{article}

\usepackage{times}
\usepackage{epsfig}
\usepackage{graphicx}
\usepackage{amsmath}
\usepackage{amssymb}
\usepackage{kotex}
\usepackage{booktabs}
\usepackage{array}
\usepackage{multirow}
\usepackage{paralist}
\usepackage{caption}
\usepackage{diagbox}
\usepackage[dvipsnames]{xcolor}
\usepackage{dsfont}
\usepackage{pifont}
\usepackage{makecell}
\usepackage[accsupp]{axessibility}

\usepackage{authblk}

\renewcommand*{\Affilfont}{\normalsize}
\makeatletter
\renewcommand\AB@affilsepx{\quad \protect\Affilfont}
\makeatother
\usepackage{cvpr}              

\definecolor{sred}{rgb}{0.8,0.0,0.0}
\definecolor{sgreen}{rgb}{0,0.8,0}

\def\gcheck{\color{sgreen}{\checkmark}}
\def\rx{\color{sred}{\ding{55}}}

\definecolor{cvprblue}{rgb}{0.21,0.49,0.74}
\usepackage[pagebackref,breaklinks,colorlinks,allcolors=cvprblue]{hyperref}

\def\paperID{11913} 
\def\confName{CVPR}
\def\confYear{2025}

\title{PersonaBooth: Personalized Text-to-Motion Generation}

\author[1,2,3]{\mbox{Boeun Kim}}
\author[2]{\mbox{Hea In Jeong}}
\author[3]{\mbox{JungHoon Sung}}
\author[1]{\mbox{Yihua Cheng}}
\author[2]{\mbox{Jeongmin Lee}}
\author[4]{\mbox{Ju Yong Chang}}
\author[3]{\mbox{Sang-Il Choi}}
\author[3]{\mbox{Younggeun Choi}}
\author[2]{\mbox{Saim Shin}}
\author[2]{\mbox{Jungho Kim}}
\author[1]{\mbox{Hyung Jin Chang}}
\affil[1]{University of Birmingham}
\affil[2]{Korea Electronics Technology Institute}
\affil[3]{Dankook University}
\affil[4]{Kwangwoon University}
\affil[ ]{\href{http://boeun-kim.github.io/page-PersonaBooth}{http://boeun-kim.github.io/page-PersonaBooth}}

\begin{document}
\maketitle
\begin{abstract}
This paper introduces Motion Personalization, a new task that generates personalized motions aligned with text descriptions using several basic motions containing Persona.
To support this novel task, we introduce a new large-scale motion dataset called \textbf{PerMo (PersonaMotion)}, which captures the unique personas of multiple actors.
We also propose a multi-modal finetuning method of a pretrained motion diffusion model called \textbf{PersonaBooth}.
PersonaBooth addresses two main challenges: i) A significant distribution gap between the persona-focused PerMo dataset and the pretraining datasets, which lack persona-specific data, and ii) the difficulty of capturing a consistent persona from the motions vary in content (action type).
To tackle the dataset distribution gap, we introduce a persona token to accept new persona features and perform multi-modal adaptation for both text and visuals during finetuning.
To capture a consistent persona, we incorporate a contrastive learning technique to enhance intra-cohesion among samples with the same persona.
Furthermore, we introduce a context-aware fusion mechanism to maximize the integration of persona cues from multiple input motions.
PersonaBooth outperforms state-of-the-art motion style transfer methods, establishing a new benchmark for motion personalization.
\end{abstract}    
\section{Introduction}
\label{sec:introduction}

\begin{figure}[t]
\centering
    \includegraphics[width=0.9\linewidth]{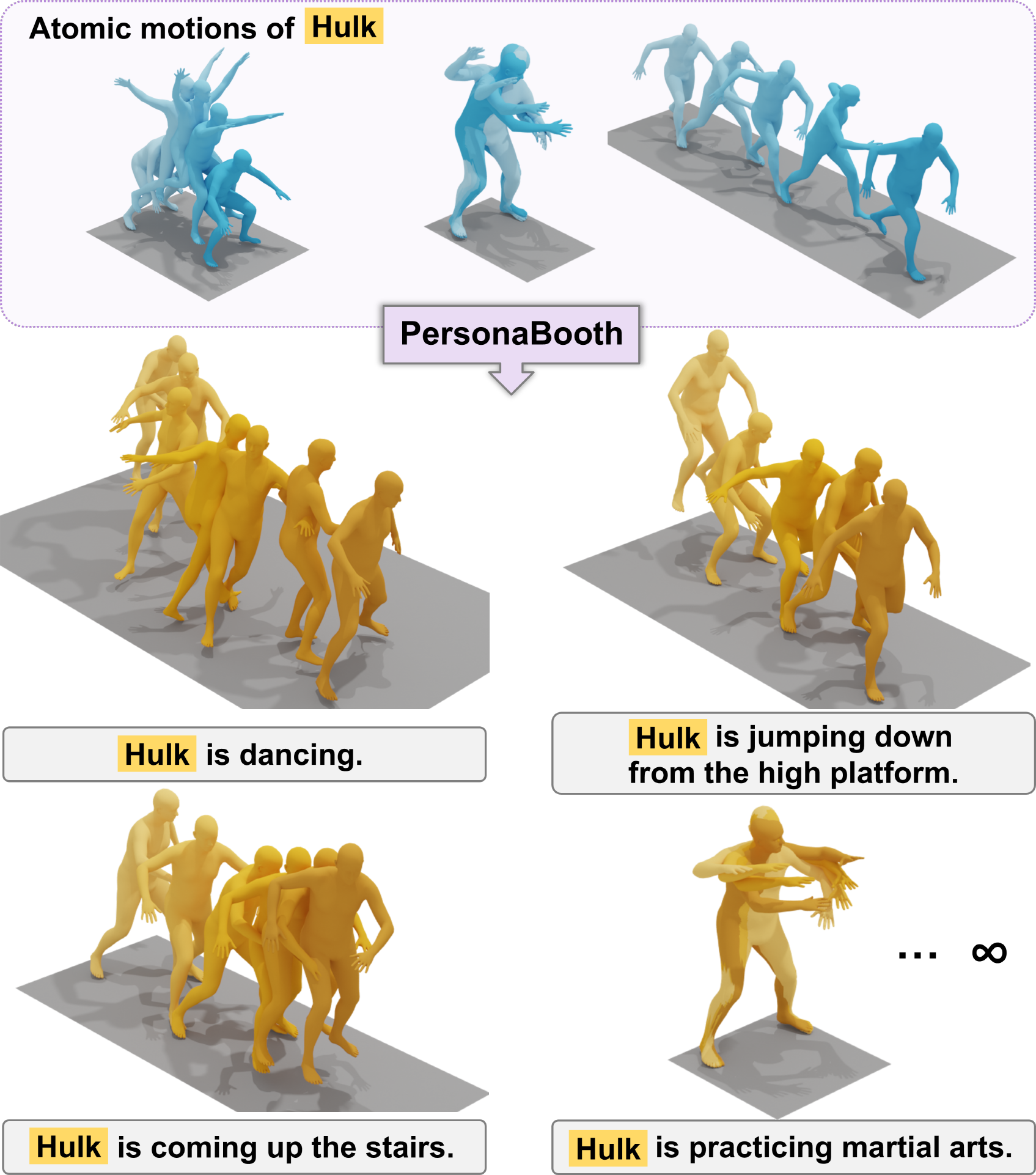}
    \caption{
    Motion Personalization generates text-driven, personalized motions based on persona embedded in atomic input motions. We propose a new framework, \textbf{PersonaBooth}, along with a new benchmark dataset, \textbf{PerMo}, for Motion Personalization
    }
    \label{fig:teaser}
\vspace{-0.3cm}
\end{figure}

Imagine a scenario where, by recording just a few basic movements of yourself, an avatar in a virtual world can mirror your personal traits and unique motion style. This kind of `motion personalization' enables realistic interactions in virtual spaces such as games and the metaverse~\cite{lee2002interactive}.
Moreover, if avatar motions could be directed through text, creating video content would become easy without the need for real-life stunt actors~\cite{videoworldsimulators2024, polyak2024movie}.

In this paper, we propose a task called \textit{Motion Personalization}, which aims to generate text-based motions that reflect individual personas using a few key atomic movements, such as jumping, punching, or walking.
We define a persona as the unique style expression of an individual.
As illustrated in Fig.~\ref{fig:teaser}, this task aims to generate a broad range of realistic motions robustly.
To facilitate this, we introduce a model called \textbf{PersonaBooth} and a new benchmark, \textbf{PerMo (PersonaMotion)}, specifically designed for motion personalization. 
Table~\ref{tab:task} provides a comparison of similar motion generation tasks. Text-to-Motion Generation approaches~\cite{petrovich2022temos, guo2022generating} generate motion using only text as input, without any reference motion, and numerous diffusion-based methods~\cite{zhang2022motiondiffuse, chen2023executing, tevet2023human} have been proposed for this task. Motion Style Transfer (MST) focuses on transferring style from a single source motion. While several datasets and methods are available for MST~\cite{raab2024monkey, zhong2024smoodi, song2024arbitrary, xia2015realtime}, the use of a single source motion restricts the ability to generate the broad and diverse motions seen in real life~\cite{kim2024most}. The proposed \textit{Motion Personalization} has no existing prior approaches or datasets. Furthermore, our proposed method introduces multi-modal finetuning—a process neglected in existing diffusion-based MST methods.

\begin{table}[t]
\caption{Comparison of motion generation tasks}
\label{tab:task}
\vspace{-0.3cm}
\setlength\tabcolsep{4pt}
\renewcommand\arraystretch{1}
\resizebox{\linewidth}{!}{%
\centering
\begin{tabular}{l|ccc}
\toprule[1pt]
Motion Task  & \makecell{Existing \\ Datasets} & \makecell{Source\\Motions} & \makecell{Multi-modal\\Finetuning}\\ \midrule
T2M Generation~\cite{zhang2022motiondiffuse, chen2023executing, tevet2023human}&
\cite{guo2022generating} \cite{plappert2016kit} &
None &
\rx
\\
Style Transfer~\cite{raab2024monkey, zhong2024smoodi, song2024arbitrary}  &
\cite{xia2015realtime}\cite{aberman2020unpaired}\cite{mason2022real} &
Single &
\rx
\\ \midrule
Personalization (Ours) &
None &
Multiple &
\gcheck
\\
\bottomrule[1pt]
\end{tabular}
}
\vspace{-0.3cm}
\end{table}


PersonaBooth is a multi-modal finetuning method for pretrained text-to-motion diffusion models. The challenge in finetuning diffusion models for the motion domain, compared to the image domain~\cite{ruiz2023dreambooth, gal2022image, zhang2024survey, li2024blip, ma2024subject, shi2024instantbooth}, lies in the relatively lower diversity of pretraining data. The most commonly used pretraining dataset, HumanML3D~\cite{guo2022generating}, contains only 15K samples, which is significantly smaller than the large-scale image datasets with 400M-650M samples~\cite{rombach2022high, ramesh2022hierarchical}. Additionally, HumanML3D includes few data reflecting personality traits, creating a substantial distribution gap when finetuning with the persona-focused PerMo dataset. 
MoMo~\cite{raab2024monkey} found that only 55 samples include styles for `run,' `walk,' `jump,' and `dance' locomotions in HumanML3D, and we could not find any persona-related samples.
Existing diffusion-based MST methods~\cite{raab2024monkey, zhong2024smoodi} consider only the visual features of the input motion and rely on a fixed text feature, which limits their ability to incorporate new data into the textual component. This inflexibility can lead to a phenomenon known as `forgetting', where the pretrained model loses its initial capabilities when finetuned with new data~\cite{zhong2024smoodi}. To address this, we introduce learnable \textit{persona tokens} to capture persona features from new data and propose an adaptation scheme for both text and visuals.

Another challenge is extracting a consistent persona across different atomic motions, which manifests in diverse ways depending on the action content. For example, the elegance of a ballerina is expressed in her feet as she walks and in her hands as she waves.
MST methods have faced similar challenges in disrupting the content of the target motion when transferring styles from the source motion with different content. Approaches like those of Park \etal~\cite{park2021diverse}, Jang \etal~\cite{jang2022motion}, and MoMo~\cite{raab2024monkey} cite this as a limitation. In contrast, MoST~\cite{kim2024most} addresses the issue with a style disentanglement loss, while MCM-LDM~\cite{song2024arbitrary} uses the target motion’s trajectory to preserve content. However, MoST’s loss function requires running the model twice for two different input motions, which is computationally expensive, especially for diffusion models with numerous steps. Additionally, since our goal is to generate motion from text rather than from target motions, the method used by MCM-LDM is not applicable. 
Therefore, we introduce a novel contrastive learning-based loss specifically designed for text-to-motion diffusion, called \textit{persona cohesion loss}. This loss facilitates cohesion across motion features with different content but the same persona.

In addition, while MST uses a single input motion, we propose a new fusion method for multiple motions. In the image domain, InstantBooth~\cite{shi2024instantbooth} simply averages persona features, but this leads to unnatural blending in the motion. To address this, we introduce \textit{Context-Aware Fusion (CAF)}, which assigns weights to the fusion of persona features based on the similarity between the prompt and input motions.
The summarized contributions of this paper are:
\begin{compactitem}
    \item We introduce the Motion Personalization task and present PerMo, a new large-scale persona dataset captured by multiple actors.
    \item We propose PersonaBooth, a multi-modal finetuning method for diffusion models that introduces text adaptation through a persona token.
    PersonaBooth applies contrastive learning to effectively capture personas across various content types.
    \item PersonaBooth achieves strong performance in both Motion Personalization and MST tasks and maximizes performance through a CAF for multiple inputs.
\end{compactitem}

\section{Related Work}
\label{sec:relatedwork}


\begin{figure*}[t]
\centering
    \includegraphics[width=\linewidth]{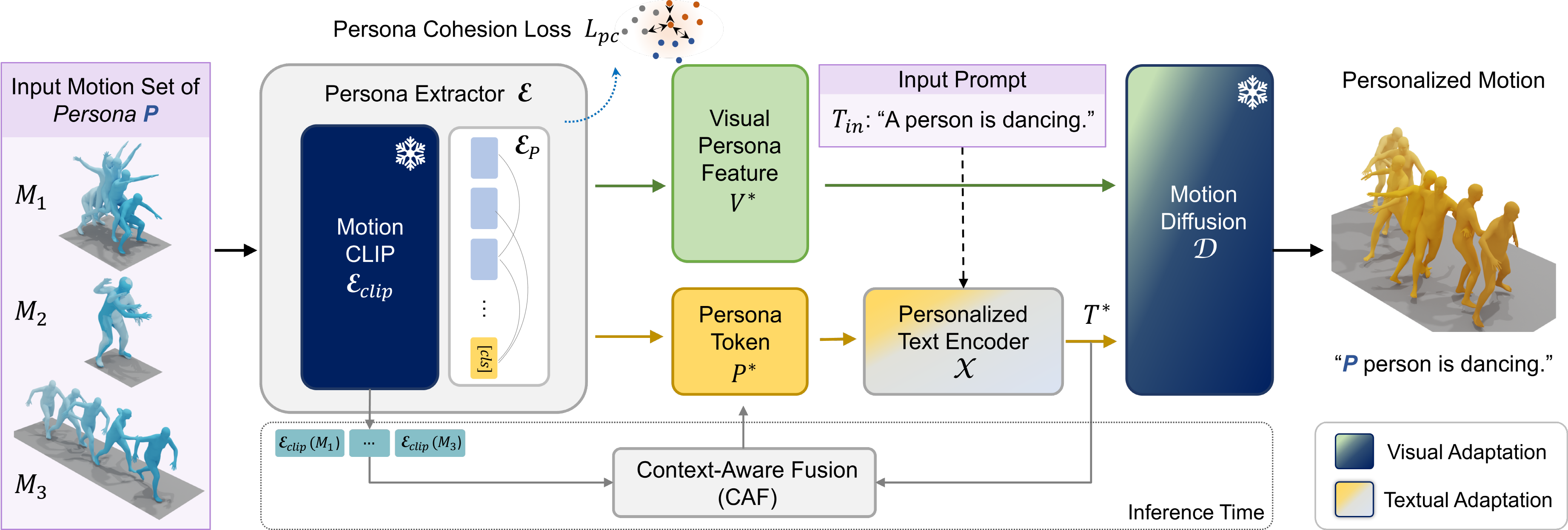}
    \caption{
    The overall framework of PersonaBooth. PersonaBooth has two adaptation paths—visual and text—for finetuning the Motion Diffusion model ($\mathcal{D}$). The Persona Extractor extracts both a visual persona feature ($V^*$) and a persona token ($P^*$) from the input motions. $V^*$ is input into the adaptive layer of $\mathcal{D}$, while $P^*$ is processed together with the input prompt through a Personalized Text Encoder, generating a personalized text feature, which is then input to $\mathcal{D}$. The entire model is trained with a classifier-free approach, incorporating a Persona Cohesion Loss. During inference, Context-Aware Fusion is applied for multiple input cases.}
    \label{fig:framework}
\vspace{-0.3cm}
\end{figure*}
The Motion Style Transfer (MST) aims to transfer the style from a source motion to a target motion.
The pioneering work by Aberman \etal~\cite{aberman2020unpaired} introduced the AdaIN layer, which simply replaces style statistics in specific layers during training. MotionPuzzle~\cite{jang2022motion} proposes a Graph Convolution Network-based framework that transfers manually specified motion to a desired body part, enabling the combination of multiple source motions.
MoST~\cite{kim2024most} addresses the issue of unwanted blending between different motion contents by introducing a transformer-based model~\cite{kim2022global} that more clearly separates style and content.

Recent MST algorithms have seen notable performance gains through the use of diffusion models. 
MoMo~\cite{raab2024monkey} proposes a zero-shot motion transfer method using a pretrained diffusion model, where it mixes two input motions within the attention module of the diffusion model. However, MoMo identifies a limitation: when the source motion content differs significantly from the target motion or text, the generated motion may not follow the target accurately.
For example, applying the style of a stationary source motion to a walking target motion may still result in a stationary output.
SMooDi~\cite{zhong2024smoodi} proposes a finetuning method for a pretrained diffusion model.
However, due to the forgetting issue, it retrained the model with both the original pretraining dataset and the style dataset. Additionally, the use of classifier-based training and the adaptation approach of creating a trainable copy of the entire model led to a training speed 10 times slower than the base diffusion model.
MCM-LDM~\cite{song2024arbitrary} addresses the challenges regarding two different contents of input motions by explicitly extracting the trajectory from the target motion and using it as a condition for the diffusion model. However, it faces a limitation: only trajectories similar to the target motion are generated, which reduces diversity. This contradicts our goal of creating diverse motions based on text prompts.


\section{Methodology}
\label{sec:method}

We propose a method for multi-modal finetuning of a text-to-motion diffusion model, named PersonaBooth. 
The overall framework is shown in Fig.~\ref{fig:framework}, and we aim to finetune the pretrained motion diffusion model, $\mathcal{D}$.
The input includes multiple atomic motions $\{M_i\}$ and a text description $T_{in}$. PersonaBooth generates personalized motion that reflects the description.
We use the same motion representation as HumanML3D~\cite{guo2022generating}, $M_i \in \mathbb{R}^{f \cdot 263}$, where $f$ is the number of frames, and 263 includes joint rotations, positions, velocities, and foot contact.
First, the Persona Extractor $\mathcal{E}$ processes each input motion to extract a visual persona feature $V^*$ and a \textit{persona token} $P^*$.
Built on a pretrained motion clip model~\cite{petrovich2023tmr} ($\mathcal{E}_{clip}$), which captures the general characteristics of motion sequences, $\mathcal{E}$ includes a transformer structure ($\mathcal{E}_P$) specially designed to extract persona features.
We applied a contrastive learning scheme to enhance the persona extraction capability of $\mathcal{E}_P$, which will be detailed in Sec.~\ref{sec:persona_extractor}.
The extracted $V^*$ is then fed into a newly added adaptive layer in $\mathcal{D}$.
Meanwhile, the Personalized Text Encoder $\mathcal{X}$ adaptively incorporates $P^*$ to produce a personalized text feature $T^*$.
$T^*$ is input into $\mathcal{D}$, where $\mathcal{D}$ ultimately generates the personalized motion.
During training, a single motion-text pair is provided to enable reconstruction. 
During inference, given motion sets $\{M_i\}$, the Context-Aware Fusion module generates weights to perform a weighted combination of the elements within $\{V^*_i\}$ and $\{P^*_i\}$.
PersonaBooth explores efficient training by using a classifier-free guidance approach.
During training, $\mathcal{E}_{clip}$ and $\mathcal{D}$, excluding the newly added adaptive layer, are kept frozen.
Sec.\ref{sec:persona_extractor} to Sec.\ref{sec:training} describe the process based on the training of single motion, $M$. The inference process for multiple input motions $\{M_i\}$ is explained in Sec.\ref{sec:CAF}.

\subsection{Persona Extractor}
\label{sec:persona_extractor}
Persona Extractor $\mathcal{E}$ extracts the \textit{persona token} $P^*$ along with the \textit{visual persona feature} $V^*$.
We used the TMR~\cite{petrovich2023tmr} structure for $\mathcal{E}_{clip}$, replacing their text encoder with the CLIP text encoder~\cite{radford2021learning} and retraining it. The CLIP text encoder is the same one used in the Personalized Text Encoder $\mathcal{X}$, and therefore, this replacement enables $P^*$ to share the same token embedding space as the input prompt.
For input motion $M$, persona features $V^*$ and $P^*$ are extracted as
\begin{align}
    V^* &= \mathcal{E}({[cls], M]}), \\
    P^* &= \mathrm{MLP}(Y), \; \text{where} \;\; Y = V^*[0],
 \end{align}
where $[cls]$ denotes a class token.

To ensure the coherence of persona features across different motion contents of the same persona, we introduce the \textit{persona cohesion loss} $L_{pc}$. We utilize a supervised contrastive learning scheme~\cite{khosla2020supervised}, where motion samples from the same persona are encouraged to be closer in the persona feature space, while motion samples from different personas are pushed apart. This helps to guide the persona features to form well-defined clusters for each persona.
For a batch containing $N$ input motion data, $L_{pc}$ is applied to a total of $2N$ data points, with one positive sample for each input motion. The positive sample is randomly selected from the same persona group as the input motion.
$L_{pc}$ for a positive pair of index $(i, j)$ is defined as
\begin{equation}
    \fontsize{9.5}{11}\selectfont
    L_{pc} = - \log \frac{\exp(\mathrm{sim}(h(Y_i), h(Y_j)) / \tau)}{\sum_{k=1}^{2N} \mathds{1}_{[k \neq i]} \exp(\mathrm{sim}(h(Y_i), h(Y_k)) / \tau)},
\end{equation}
where $h(\cdot)$ denotes projection head. 
$Y_j$ is extracted from $M_j$ which belongs to the same persona group as $M$.
$\text{sim}(u,v)$  indicates the cosine similarity between two vectors 
$u$ and $v$. $\tau$ denotes the temperature parameter.

\subsection{Textual and Visual Adaptation}
\label{sec:ref}
\begin{figure}[t]
\centering
    \includegraphics[width=\linewidth]{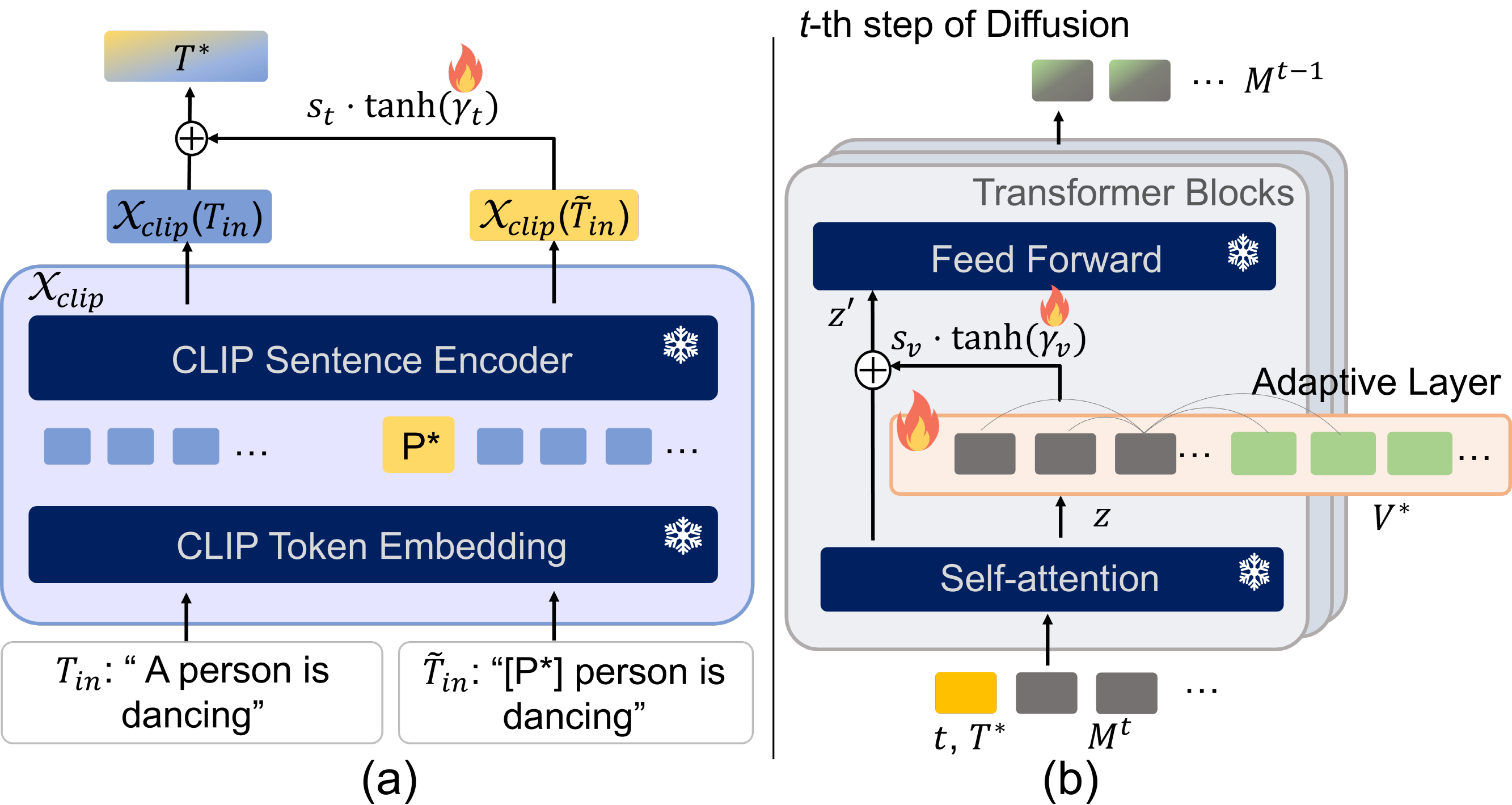}
    \vspace{-0.6cm}
    \caption{ Textual and visual adaptation. (a) Personalized Text Encoder, $\mathcal{X}$. (b) $t$-th step of the Motion Diffusion, $\mathcal{D}$.
    Learnable parameters are denoted by the fire icon
    }
    \label{fig:adapt}
\vspace{-0.5cm}
\end{figure}
The discrepancy between our PerMo dataset and the non-persona-specific pretraining dataset, HumanML3D~\cite{guo2022generating} necessitates a carefully designed adaptation mechanism for effective finetuning.
To address this, we propose two adaptation paths—textual and visual—to seamlessly integrate the two persona features, $P^*$ and $V^*$.

The structure of the Personalized Text Encoder ($\mathcal{X}$) is shown in Fig.~\ref{fig:adapt} (a).
Given an input sentence like ``A person is dancing" ($T_{in}$), we introduce a personalized sentence ``[P*] person is dancing" ($\tilde{T}_{in}$), where the word [P*] precedes the subject.
Subsequently, the vector $P^*$, derived from $\mathcal{E}$, replaces the token embedding corresponding to [P*].
Then, the modified sentence embedding, $\mathcal{X}_{clip}(\tilde{T}_{in})$, is adaptively merged with the original sentence embedding, $\mathcal{X}_{clip}(T_{in})$.
The descriptions in HumanML3D are generally structured with subjects like `A Person,' or `Someone' and lack adjectives to characterize the subject. Thus, a smoother adaptation is necessary to enable the pretrained model to handle variations in sentence structure effectively.
Specifically, personalized text feature $T^*$ is derived as
\begin{align}
\label{eq:text_adapt}
    T^* &= \mathcal{X}(T_{in}, P^*) \\
     &= \mathcal{X}_{clip}(T_{in}) + s_t \cdot \tanh(\gamma_t) \cdot \mathcal{X}_{clip}(\tilde{T}_{in}, P^*), \notag
\end{align}
where $\gamma_t$ is a learnable parameter initialized as zero which is for zero gating~\cite{zhang2023llama}. 
$s_t$ is a scaling factor~\cite{shi2024instantbooth}, a constant used to balance the adaptive layer during inference.
CLIP text encoder $\mathcal{X}_{clip}$ is frozen during training.

For visual adaptation, we introduce a single adaptive layer into the transformer structure, following the approach used in~\cite{shi2024instantbooth} for the UNet.
As illustrated in Fig.~\ref{fig:adapt} (b), for $b$-th transformer block in $t$-th step of the diffusion, the adaptive layer is placed between the original self-attention layer and the feedforward layer as
\begin{align}
    z' = z + s_v \cdot \tanh(\gamma_v) \cdot \mathrm{Adapt}([z, V^*]),
    \label{eq:visual_adapt}
\end{align}
where $\mathrm{Adapt}$ indicates the adaptive layer and $z$ and $z'$ denote the input and output features of the adaptive layer, respectively.
$\gamma_v$ and $s_v$ serve as the gating parameter and scaling factor.
We evaluated several adaptive layer structures, including self-attention, cross-attention, and the AdaIN layer, which is widely used in style transfer~\cite{aberman2020unpaired, kim2024most}. Among these, self-attention was found to be the most effective.

\subsection{Training}
\label{sec:training}
We used a pretrained 50-step MDM~\cite{tevet2023human} for $\mathcal{D}$. 
Diffusion is modeled as a Markov noising process $\{M^t\}_{t=0}^T$, where $M^0$ is sampled from the training motion, and $t$ denotes the noise time-step.
The diffusion loss for finetuning is given by
\begin{equation}
    \fontsize{9.5}{11}\selectfont
    L_{D} := \mathbb{E}_{M^0, t,T} \left[ \left\| M^0 - \mathcal{D} \left( M^t, t, V^*, T^* \right) \right\|_2^2 \right] + L_{geo},
    \end{equation}
where $L_{geo}$ is the geometric loss used during the MDM pretraining~\cite{tevet2023human}. The final training loss is expressed as
\begin{align}
    L = L_{D} + \lambda L_{pc},
\end{align}
where $\lambda$ is a weight for the $L_{pc}$.

We utilize Classifier-Free Guidance (CFG)~\cite{ho2022classifier}, which is a technique for adjusting the trade-off between the diversity and fidelity of generated samples. Similarly, we applied CFG during finetuning to enable control over the trade-off between the pretrained diffusion model’s capability for generating diverse motions and fidelity to the newly provided conditions, $V^*$ and $T^*$.
During training, $V^*$ and $T^*$ are randomly dropped with a 10\% probability, respectively. 
During sampling, the output of $t$-th step is extrapolated as
\begin{align}
    &\mathcal{\hat{D}}(M^t, t, V^*, T^*) = b \mathcal{D}_T + (1-b) \mathcal{D}_V, \\
    &\resizebox{\linewidth}{!}{$\mathcal{D}_T = \mathcal{D}(M^t, t, V^*, \emptyset)  + g_t (\mathcal{D}(M^t, t, V^*, T^*) - \mathcal{D}(M^t, t, V^*, \emptyset)),$} \notag\\
    &\resizebox{\linewidth}{!}{$\mathcal{D}_V = \mathcal{D}(M^t, t, \emptyset, T^*)  + g_v (\mathcal{D}(M^t, t, V^*, T^*) - \mathcal{D}(M^t, t, \emptyset, T^*)),$} \notag
\end{align}
where $\emptyset$ indicates the absence of an input.
$g_t$ and $g_v$ represent guidance scales, while $b$ serves as the balancing factor between the modalities. 
As $g_v$ increases, the model puts more focus on the persona inherent in the visual feature.
However, if $g_v$ is set too high, inconsistencies with the text prompt may arise. 
On the other hand, $g_t$ controls the model's adherence to the prompt, with the intensity of $P^*$ in the text being adjusted by $s_t$ in Eq.~(\ref{eq:text_adapt}). 
The hyperparameter settings are specified in Sec.~\ref{sec:experiments}, and an ablation study for these is provided in the supplementary material.

\begin{figure}[t]
\centering
    \includegraphics[width=\linewidth]{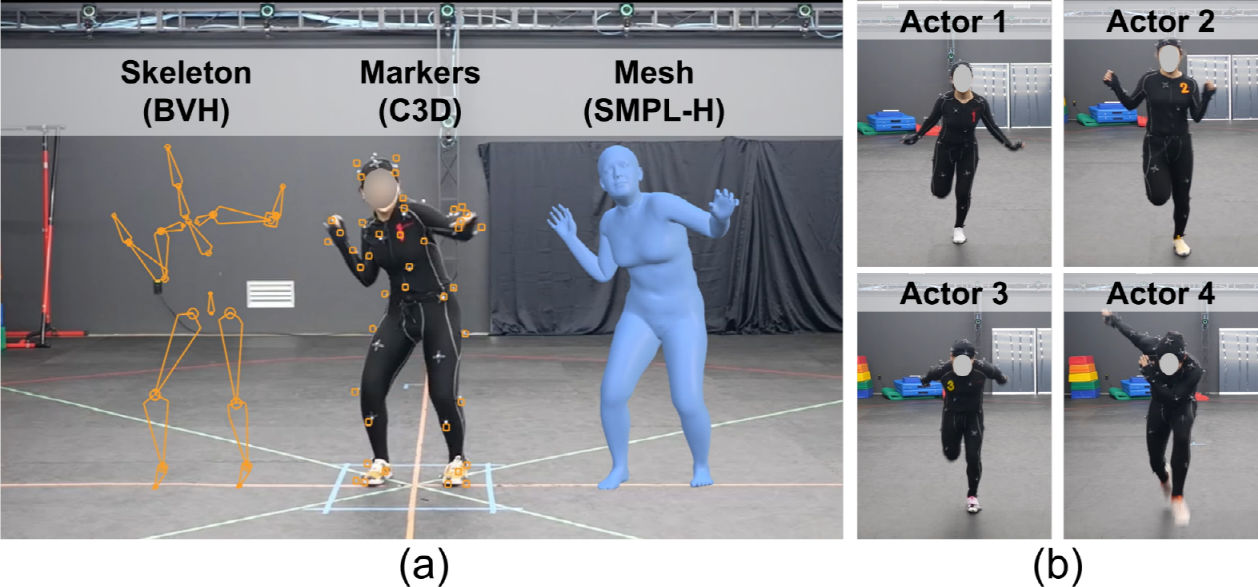}
    \includegraphics[width=\linewidth]{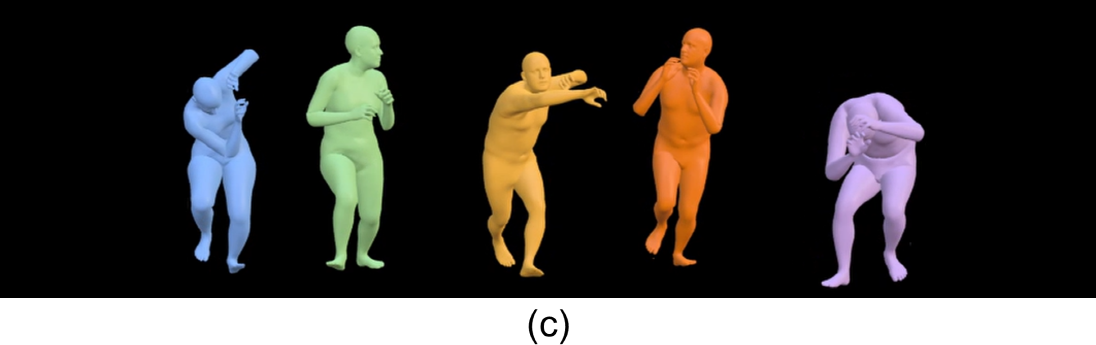}
    \vspace{-0.7cm}
    \caption{(a) Motion capture studio and examples of data formats: skeleton, markers, and mesh.
    (b) Unique persona expressions of each actor in the `Childish' category. (c) Rendered mesh for each actor in the `Fearful' category
    }
    \label{fig:dataset}
\vspace{-0.5cm}
\end{figure}

\subsection{Context-Aware Fusion for Multiple Inputs}
\label{sec:CAF}
Intuitively, if multiple motions are input, these additional cues can help with generating motion.
However, it's also crucial to choose the right cues, as transferring persona from motions with very different content types, even with $L_{pc}$, can result in unnatural movements. A straightforward approach might be to take the mean of all features, as suggested in~\cite{shi2024instantbooth}. In the motion domain, however, taking all motions can result in blended motions that appear implausible.
To overcome this, we introduce a Context-Aware Fusion (CAF) method that prioritizes input motions based on their contextual relevance to the input prompt. To assess this relevance, we use the motion encoder and text encoder of the motion clip ($\mathcal{E}{clip}$ and $\mathcal{X}{clip}$), which align the text and motion feature spaces.
The CAF is formulated as
\begin{align}
    I_{\text{Top-k}} &= \text{argmax}_i(S_i), \quad S_i = \text{sim} (\mathcal{E}_{\text{clip}}(M_i), \bar{T}^{*}), \\
    w_i &= 
    \begin{cases} 
    \dfrac{\exp(S_i)}{\sum_n \exp(S_n)}, & \text{if } i \in I_{\text{Top-k}}, \\
    0, & \text{otherwise},
    \end{cases}
\end{align}
where $\bar{T}^*$ represents the personalized text feature, using the average value of $\{P^*\}$ as a persona token. sim$(\cdot)$ denotes cosine similarity.
Consequently,
\begin{align}
V^* = \sum_i w_i V_i^*, \quad P^* = \sum_i w_i P_i^*.
\end{align}

\section{PerMo Dataset}
\label{sec:dataset}

We collected a large-scale PerMo dataset, capturing personas from multiple actors. To ensure variety, we hired five professional motion capture actors of diverse genders and body types. Each actor is assigned to perform 34 styles, categorized into \textit{Age, Character, Condition, Emotion, Traits}, and \textit{Surroundings}, resulting in a total of 170 personas.
Each actor performed 10 distinct contents for every style, carefully selected to engage different body parts.

\begin{table}[t]
\caption{Comparison with existing motion style datasets. The note for `*' is provided in the main text. `-' indicates no information}
\label{tab:dataset}
\setlength\tabcolsep{3pt}
\centering
\small
\begin{tabular}{l|c|c|c|c|c|c}
\toprule[1pt]
Dataset & Actors & Styles & Contents & Clips & Mesh & Text\\ \midrule
    Xia \cite{xia2015realtime}&
    - &
    8 &
    6 &
    572 &
    \rx &
    \rx \\
    BFA \cite{aberman2020unpaired}&
    1 &
    16 &
    9 &
    32 &
    \rx &
    \rx  \\
    BN-1 \cite{kobayashi2023motion}&
    2* &
    15 &
    17 &
    175 &
    \rx &
    \rx  \\
    BN-2 \cite{kobayashi2023motion}&
    2* &
    7 &
    10 &
    2,902 &
    \rx &
    \rx  \\
    100Style \cite{mason2022real}&
    1 &
    100* &
    8* &
    810 &
    \rx &
    \rx  \\ \midrule
    \textbf{PerMo}(ours) &
     \color{red}{\textbf{5}} &
    34 &
    10 &
     \color{red}{\textbf{6,610}} &
    \gcheck &
    \gcheck\\
\bottomrule[1pt]    
\end{tabular}
\vspace{-0.3cm}
\end{table}

The dataset was captured in a studio equipped with 33 OptiTrack cameras, using 41 optical markers per person. For compatibility with other motion datasets, we converted the marker data to SMPL-H~\cite{romero2022embodied} format using Mosh++~\cite{mahmood2019amass}. All data underwent expert cleaning and verification, with the entire process adhering to strict quality guidelines.
Fig.~\ref{fig:dataset} (a) illustrates the motion capture studio and the provided data formats, while (b) highlights the distinct personas expressed by each actor.
Additionally, we constructed 20-30 varied descriptions for each content type, following research indicating that high-quality finetuning descriptions improve prompt-based control~\cite{he2023data}. Starting with one detailed description, we instructed ChatGPT to generate a range from high-level descriptions (e.g., “A person is jumping ahead”) to low-level descriptions (e.g., “A person pushes off the ground and jumps forward several times”).
Additional details on data categories, capture process, data structure, and examples can be found in the supplementary.

\begin{figure*}[t]
\centering
    \includegraphics[width=\linewidth]{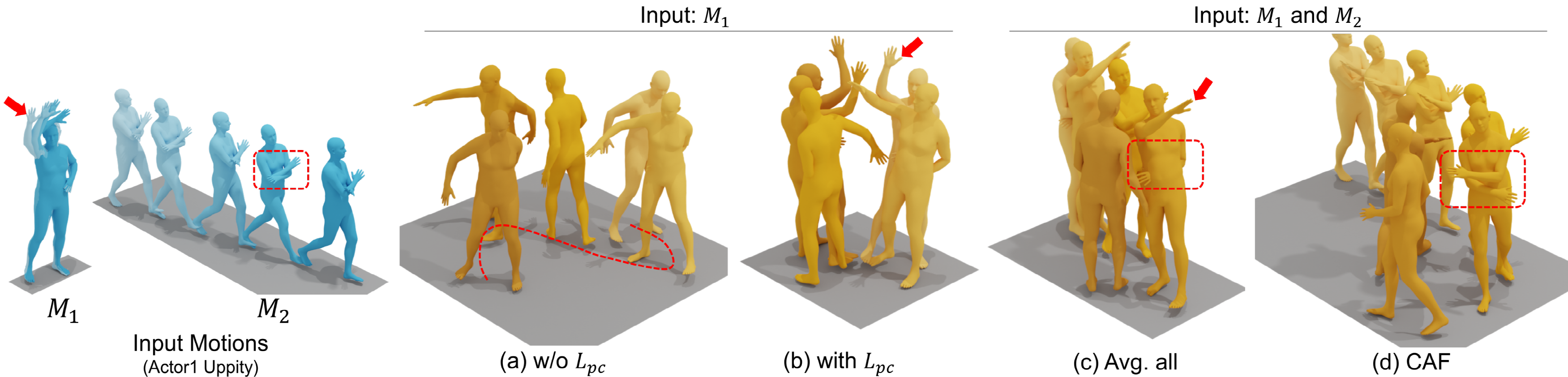}
    \caption{
    Example of the ablation study. The input motions are from the `Uppity' of Actor 1. The input prompt is ``A person walks in a circle." In (a) and (b), only $M_1$ is provided for the input, while both $M_1$ and $M_2$ are provided for (c) and (d). $L_{pc}$ encourages the generated motion to closely follow the prompt, while CAF prevents the motion from blending. We set $k=1$ for CAF
    }
    \label{fig:ablation}
\end{figure*}

Table~\ref{tab:dataset} shows a comparison with existing motion style datasets. 
Notably, PerMo is the first dataset to collect data from multiple actors. Our dataset also offers the highest number of total clips and content categories among trainable datasets and is the only one that includes mesh, marker data, and descriptions.
The BN datasets~\cite{kobayashi2023motion} involve two actors; however, for the same category, only one actor is used (BN-1), or there are no tags identifying the actors (BN-2). 
BN-1 is unsuitable for training, as it contains only a single motion sequence per category.
The 100Style~\cite{mason2022real} offers the largest variety of styles, but 58 of its categories focus on content, rather than on style~\cite{zhong2024smoodi}.
Additionally, the eight content categories in this dataset mainly consist of walking or running, offering limited variation.


\section{Experiments}
\label{sec:experiments}
\textbf{Implementation Details.} 
We conduct experiments using the proposed PerMo and 100Style~\cite{mason2022real} datasets. During training, a single motion is cropped differently to serve as both the input and the GT motion for generation.
We set $\lambda$ to $10^{-2}$ and use the AdamW~\cite{loshchilov2017decoupled} optimizer with a learning rate of $10^{-4}$. Training runs for 500 epochs with a batch size of 64. We set $s_t$ and $s_v$ to 1 during training and to 0.3 during inference. $g_t$ is set to 10, and $g_v$ to 15 and 10 for PerMo and 100Style, respectively. We set $b$ to 0.7 for single-input and 0.5 for multi-input settings, and $k$ to 5 for CAF.
\\
\textbf{Evaluation Settings.} 
For quantitative evaluation, we use descriptions from the HumanML3D test set as prompts and motion samples from the PerMo and 100Style datasets as inputs.
For each description, $[P^*]$ is logically placed as a modifier for the subject.
We used two evaluation settings: Single Input (SI) and Multiple Input (MI).
In SI, one motion is sampled from the entire dataset for comparison with MST methods.In MI, a persona set is first sampled, and then $|M_i|$ motions are drawn from it.

Performance is evaluated using Frechet Inception Distance (FID), R-Precision, and Diversity—metrics commonly used in motion generation tasks~\cite{guo2020action2motion, guo2022generating, chen2023executing, tevet2023human}—as well as Persona Recognition Accuracy (PRA), a metric adopted by Style Recognition Accuracy (SRA) in MST~\cite{jang2022motion, song2024arbitrary}.
\textbf{FID} assesses the overall quality and realism of generated motions and is our primary metric~\cite{guo2020action2motion}. \textbf{R-Precision} evaluates text-to-motion alignment accuracy~\cite{guo2022generating}, and \textbf{Diversity}~\cite{guo2020action2motion} measures how well the generated motions reflect a broad distribution of personas. 
\textbf{PRA} uses a pretrained persona classifier to evaluate the persona consistency of generated motions.
The details are in the supplementary material.
\begin{table}[t]

\caption{Ablation study of the proposed components. 
$|M_i|$=max indicates that all motions in the chosen persona set are input}
\label{tab:abl}

\resizebox{\linewidth}{!}{%
\begin{tabular}{@{}lcccccc@{}}
\toprule[1pt]
\multirow{2}{*}{Methods} & \multirow{2}{*}{FID $\downarrow$} &\multicolumn{3}{c}{R Precision $\uparrow$} & \multirow{2}{*}{PRA avg.$\uparrow$} & \multirow{2}{*}{Diversity $\uparrow$} \\ \cmidrule(lr){3-5}
 & & \multicolumn{1}{c}{Top 1} & \multicolumn{1}{c}{Top 2} & \multicolumn{1}{c}{Top 3} & & \\\midrule 
 \multicolumn{7}{c}{Single Input (SI) Setting} \\\midrule 
  baseline &
  7.45 &
  0.06 &
  0.11 &
  0.16 &
  17.99 &
  7.48
\\
   \; \; + $P^*$  &
   5.06 &
   0.05 &
   0.10 &
   0.15 &
   \textbf{18.26} &
   \textbf{8.01} 
\\
   \; \;\quad +$L_{pc}$ &
   \textbf{3.18} &
   \textbf{0.15} &
   \textbf{0.26} &
   \textbf{0.33} &
   18.05 &
   7.74
\\\midrule
\multicolumn{7}{c}{Multiple Input (MI) Setting} \\\midrule
  $|M_i|$=$1$ &
  4.27 &
  0.08 &
  0.15 &
  0.20 &
  16.20 &
  7.71
\\
  $|M_i|$=max, \cite{shi2024instantbooth} &
  3.52 &
  \textbf{0.19} &
  \textbf{0.30} &
  0.38 &
  \textbf{19.24} &
  7.88
\\
  $|M_i|$=max, CAF &
  \textbf{2.95} &
  \textbf{0.19} &
  \textbf{0.30} &
  \textbf{0.39} &
  18.13 &
  \textbf{8.12}
\\ \bottomrule[1pt]
\end{tabular}%
}
\end{table}


\subsection{Ablation Study}
Table~\ref{tab:abl} demonstrates the effects of our key contributions.  
The baseline represents results obtained using only visual adaptation. When we introduce $P^*$ for textual adaptation, the FID score drops significantly, and both PRA and Diversity increase.
This indicates that multimodal integration of persona information not only strengthens the reflection of persona but also supports the generation of plausible motion.
Furthermore, adding $L_{pc}$ leads to an additional reduction in FID and a notable increase in R-precision. 
This implies that $L_{pc}$ helps the Persona Extractor effectively capture persona essence disentangled from the input content, resulting in generated motion that aligns well with the prompt without compromising content.

\begin{figure*}[t]
\centering
    \includegraphics[width=\linewidth]{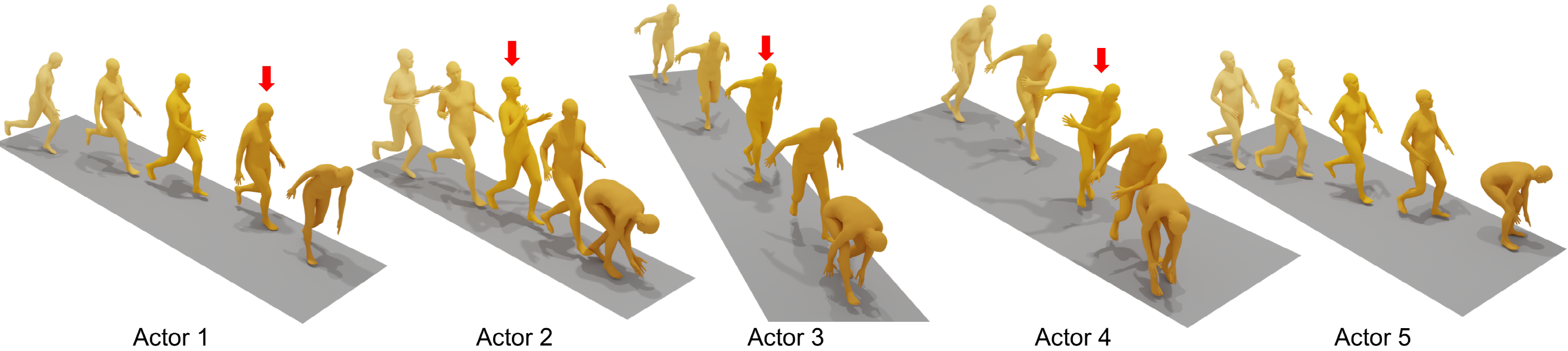}
    \caption{
    Example reflecting the different personas of Actors 1-5 for the same `Childish' category. 
    The input prompt is ``A person is running and bending to pick something." Red arrows highlight the frames that clearly show distinct movements. The input motion of each actor can be referenced in Fig.~\ref{fig:dataset} (b)
    }
    \label{fig:result}
\end{figure*}
Next, the effectiveness of CAF is validated in the MI setting. When multiple input motions are used ($|M_i|$=max), all metrics improve compared to using a single input ($|M_i|$=1) due to the higher likelihood of finding motion candidates that align with the prompt. Additionally, even in cases with multiple inputs, selecting the contextually most relevant Top-k motions—as done by CAF—is shown to be crucial. For $|M_i|$=max, \cite{shi2024instantbooth} presents a baseline approach that uses the average feature of all inputs. 
Introducing the proposed CAF reduces FID while increasing Diversity, indicating that the generated motions are sufficiently diverse and plausible rather than oversimplified by averaging.
Although the PRA score decreases, we found through qualitative results that generating plausible motion is a more crucial factor in improving motion quality.

Fig.~\ref{fig:ablation} provides an example. When using only $M_1$ as the input motion, if $L_{pc}$ is not applied, the generated motion fails to follow the path provided in the prompt, as shown in (a). However, with $L_{pc}$, the motion better aligns with the prompt’s trajectory, though it still includes unintended hand-waving motions unrelated to the prompt.
When using multiple input motions, including $M_2$ (a walking motion with arms crossed), taking the mean of features results in a blended motion, as shown in (c). This blending causes one arm to bend while the other hangs low, creating unnatural poses.
In contrast, with CAF, $M_2$ receives a higher weight, resulting in a more natural walking motion with arms crossed, as shown in (d).

\begin{table}[t]

\caption{Ablation study regarding adaptation. PersonaBooth (SI) indicates our complete model for single inputs}
\label{tab:abl_adapt}

\resizebox{\linewidth}{!}{%
\begin{tabular}{@{}lcccccc@{}}
\toprule[1pt]
\multirow{2}{*}{Methods} & \multirow{2}{*}{FID $\downarrow$} &\multicolumn{3}{c}{R Precision $\uparrow$} & \multirow{2}{*}{PRA avg.$\uparrow$} & \multirow{2}{*}{Diversity $\uparrow$} \\ \cmidrule(lr){3-5}
 & & \multicolumn{1}{c}{Top 1} & \multicolumn{1}{c}{Top 2} & \multicolumn{1}{c}{Top 3} & & \\\midrule 
AdaIN &
  8.77 &
  0.04 &
  0.07 &
  0.10 &
  15.10 &
  7.33
\\
  Cross-Attn &
   8.96 &
   0.04 &
   0.07 &
   0.10 &
   16.37 &
   7.62
\\
  w/o T adapt. &
   3.61 &
   0.14 &
   0.23 &
   0.31 &
   17.77 &
   \textbf{7.88}
\\\midrule
  PersonaBooth (SI) &
  \textbf{3.18} &
  \textbf{0.15} &
  \textbf{0.26} &
  \textbf{0.33} &
  \textbf{18.05} &
  7.74
\\ \bottomrule[1pt]
\end{tabular}%
}
\end{table}


Table~\ref{tab:abl_adapt} presents various ablation results related to adaptation. For the visual adaptation layer, Self-Attention proved to be the most effective compared to AdaIN and Cross-Attn. Additionally, introducing text adaptation at Personalized Text Encoder improves the PRA, FID, and R-precision metrics.
Fig.~\ref{fig:result} demonstrates how PersonaBooth effectively captures and represents each actor's unique action. The red arrows specifically highlight the distinct arm movements of each persona. Additional details, including variations in body tilt and movement speed, are more clearly observable in the supplementary video materials.

\begin{table*}[t]

\caption{Comparison with the state-of-the-art methods on the PerMo dataset. 
The comparison is made in the Single Input (SI) setting as the existing methods do not support multiple inputs.
MCM-LDM* indicates the model is finetuned on the PerMo dataset.}
\label{tab:comp_PerMo}
\vspace{-0.3cm}
\center
\resizebox{\linewidth}{!}{%
\begin{tabular}{@{}llcccccccccccccccc@{}}
\toprule[1pt]
\multirow{2}{*}{Methods} & \multirow{2}{*}{Steps} & \multirow{2}{*}{FID $\downarrow$} &\multicolumn{3}{c}{R Precision $\uparrow$}  & \multicolumn{10}{c}{PRA $\uparrow$} & \multirow{2}{*}{Diversity $\uparrow$} \\ \cmidrule(lr){4-6} \cmidrule(lr){7-16}
 &  &  &
 \multicolumn{1}{c}{Top 1} &
 \multicolumn{1}{c}{Top 2} &
 \multicolumn{1}{c}{Top 3} &
 \multicolumn{1}{c}{Age} &
 \multicolumn{1}{c}{Char1} &
 \multicolumn{1}{c}{Char2} &
 \multicolumn{1}{c}{Cond1} &
 \multicolumn{1}{c}{Cond2} &
 \multicolumn{1}{c}{Emo1} &
 \multicolumn{1}{c}{Emo2} &
 \multicolumn{1}{c}{Trait} &
 \multicolumn{1}{c}{Sur} &
 \multicolumn{1}{c}{\textbf{Avg.}}
 &
 \\\midrule 
 
  MoMo~\cite{raab2024monkey}&
  100 &
  13.91 &
  0.05 &
  0.09 &
  0.13 &
  28.23 &
  10.23 &
  11.30 &
  \textbf{8.73} &
  15.97 &
  9.97 &
  14.57 &
  12.60 &
  7.93 &
  13.47 &
  6.74
\\
  MCM-LDM~\cite{song2024arbitrary}&
  1000 &
  9.16 &
  0.13 &
  0.23 &
  0.30 &
  47.38 &
  11.44 &
  10.34 &
  7.63 &
  18.90 &
  11.69 &
  17.20 &
  {15.71} &
  \textbf{9.22} &
  17.00 &
  6.70
\\
  MCM-LDM*~\cite{song2024arbitrary} &
  1000 &
  9.70 &
  0.13 &
  0.23 &
  0.30 &
  46.49 &
  12.12 &
  9.22 &
  {7.95} &
  \textbf{19.77} &
  {11.75} &
  18.05 &
  15.18 &
  7.33 &
  16.76 &
  6.76
\\\midrule
  PersonaBooth (SI) &
  50 &
  \textbf{3.18} &
  \textbf{0.15} &
  \textbf{0.26} &
  \textbf{0.33} &
  \textbf{48.00} &
  \textbf{13.67} &
  \textbf{11.97} &
  6.69 &
  18.67 &
  \textbf{14.34} &
  \textbf{19.75} &
  \textbf{17.06} &
  8.80 &
  \textbf{18.05} &
  \textbf{7.74}
\\\midrule
  PersonaBooth (MI) &
  50 &
  {2.95} &
  {0.19} &
  {0.30} &
  {0.39} &
  {51.95} &
  {14.89} &
  {12.00} &
  5.16 &
  {22.95} &
  9.44 &
  {18.14} &
  15.45 &
  {10.09} &
  {18.13} &
  {8.12}
  
\\ \bottomrule[1pt]
\end{tabular}%
}
\vspace{-0.2cm}
\end{table*}


\begin{table}[t]

\caption{Comparison with the state-of-the-art methods on the 100Style dataset. MCM-LDM* indicates the model is finetuned on the 100Style dataset}
\label{tab:comp_100style}
\resizebox{\linewidth}{!}{%
\begin{tabular}{@{}lcccccc@{}}
\toprule[1pt]
\multirow{2}{*}{Methods}  & \multirow{2}{*}{FID $\downarrow$} &\multicolumn{3}{c}{R Precision $\uparrow$} & \multirow{2}{*}{SRA $\uparrow$} & \multirow{2}{*}{Diversity $\uparrow$} \\ \cmidrule(lr){3-5}
 & & \multicolumn{1}{c}{Top 1} & \multicolumn{1}{c}{Top 2} & \multicolumn{1}{c}{Top 3} & & \\\midrule 
  MoMo~\cite{raab2024monkey}&
  5.97 &
  0.07 &
  0.10 &
  0.14 &
  60.57 &
  7.82
\\
  MCM-LDM~\cite{song2024arbitrary}&
  7.20 &
  0.18 &
  0.29 &
  0.37 &
  54.44 &
  6.76
\\
  MCM-LDM*~\cite{song2024arbitrary} &
  6.53 &
  0.17 &
  0.28 &
  0.37 &
  54.61 &
  7.12
\\\midrule
  PersonaBooth (SI) &
  \textbf{3.27} &
  \textbf{0.20} &
  \textbf{0.31} &
  \textbf{0.40} &
  \textbf{64.53} &
  \textbf{7.90}
\\ \bottomrule[1pt]
\end{tabular}%
}
\vspace{-0.2cm}
\end{table}


\begin{figure*}[htbp]
    \centering
        \begin{minipage}{\textwidth}
        \centering
        \includegraphics[width=\textwidth]{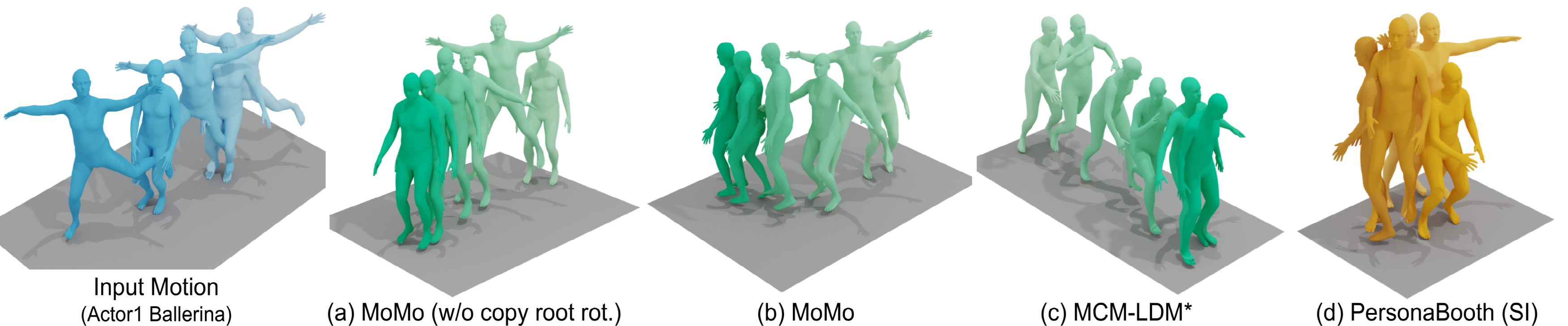}
        \vspace{-0.5cm}
        \caption{Qualitative comparison on PerMo dataset. The input prompt is ``A person hops forward and turns in the air."}
        \label{fig:comp2}
    \end{minipage}

    \begin{minipage}{\textwidth}
        \centering
        \includegraphics[width=\textwidth]{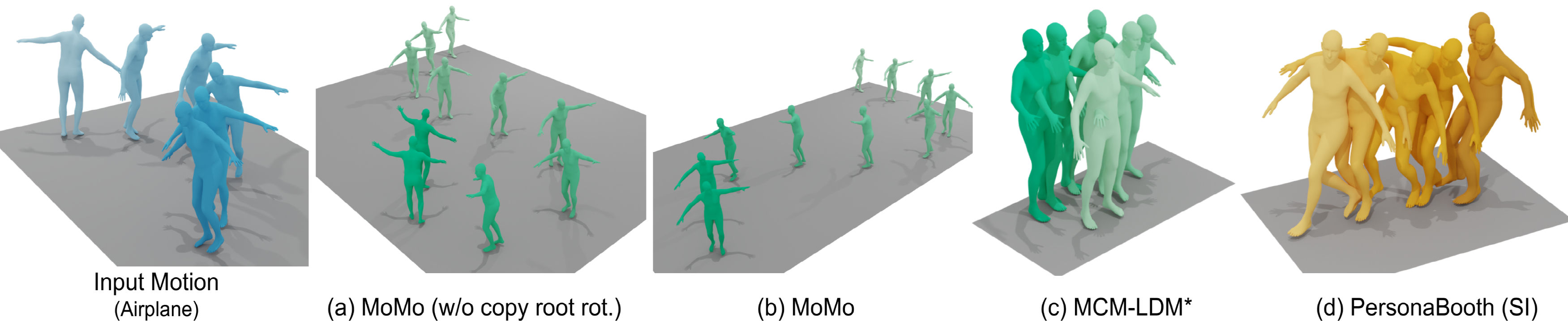}
        \vspace{-0.5cm}
        \caption{Qualitative comparison on 100Style dataset. The input prompt is ``A person walks backward and sits down on the chair." }
        \label{fig:comp1}
    \end{minipage}
    
\vspace{-0.2cm}

\end{figure*}

\subsection{Comparison with State-of-the-Art methods}
We compare our PersonaBooth with the state-of-the-art diffusion-based MST approaches, MoMo~\cite{raab2024monkey} and MCM-LDM~\cite{song2024arbitrary}, on the PerMo dataset. Additionally, we evaluate it on an existing style dataset, 100Style~\cite{mason2022real}.
As MCM-LDM does not support prompt-based generation, we used MDM-generated motions~\cite{tevet2023human} as target motions for evaluation. MCM-LDM was evaluated using the original model trained on HumanML3D, as well as finetuned models, referred to as MCM-LDM, which were trained on the PerMo and 100Style datasets. 

\subsubsection{PerMo Dataset}
The comparative results for the PerMo dataset are shown in Table~\ref{tab:comp_PerMo}, with all subcategory results displayed for PRA. The single-input PersonaBooth significantly reduces FID compared to existing methods, indicating that while previous studies struggled with PerMo—a challenging dataset containing more diverse content than 100Style—PersonaBooth performs robust in generating high-quality motion. Additionally, PersonaBooth achieves the highest performance in R-Precision, PRA, and Diversity compared to existing methods. Applying multiple inputs and CAF further enhances performance.
It’s noteworthy that while MoMo and MCM-LDM utilized a 100-step and 1000-step diffusion model, respectively, we used a 50-step diffusion model.

Fig.~\ref{fig:comp2} shows qualitative results.
MoMo~\cite{raab2024monkey} notes that their generated motions often follow the input motion's direction instead of the target's, which they attempted to copy the root rotation from the input motion (shown in (a) before and (b) after adjustment). 
The prompts for ``turns in the air" were not met in (a), (b), or (c). Particularly, (c) does not reflect the style of ballerina. In contrast, PersonaBooth accurately generates the motion by first performing a hop that reflects the style, followed by a precise turn in the air. Example videos are provided in the supplementary material.

\subsubsection{100Style Dataset}
Table~\ref{tab:comp_100style} presents the comparative results on the 100Style dataset. MoMo demonstrates high PRA and Diversity, indicating effective style reflection. However, it shows very low R-precision, suggesting poor alignment between generated motions and prompts. Finetuned MCM-LDM improved performance on nearly all metrics compared to its original model. Although finetuned MCM-LDM achieved high R-precision, it also had a high FID, resulting in unnatural motion generation, with low PRA and Diversity indicating weaker style reflection.
In contrast, our single-input PersonaBooth significantly reduces FID and achieves top performance across all metrics, including R-precision, PRA, and Diversity.

In Fig.~\ref{fig:comp1} (a) and (b), MoMo produces sliding motions that don’t align with the prompt. MCM-LDM shows sideward movements.
In contrast, PersonaBooth accurately reflects both the input motion's style, like arm extension and leaning, and the sitting motion in line with the prompt.

\section{Conclusion}
\label{sec:conclusion}
We propose the PerMo dataset and the PersonaBooth framework for Motion Personalization. Through contrastive learning, we effectively capture the essence of a persona. Furthermore, we enhance the integration of persona traits into diffusion models using a multi-modal finetuning method. We also introduce the CAF, which effectively handles multiple inputs, ensuring cohesive representation of personalized motion characteristics.
\\
\textbf{Limitations and Future Works.}
The PerMo contains relatively short motion sequences, which sometimes lead to stationary poses at the end of generated motions of the same length. However, this issue does not occur when prompted with longer actions, suggesting that automatic adjustment of motion length remains a task for future work. In addition, we plan to investigate approaches for applying CAF to each sequential action within a prompt individually.
For instance, applying CAF separately to `run' and `jump' in the prompt ``A person runs and then jumps."
\vspace{0.1cm}
\\
\textbf{Acknowledgement.}
\\
This work was supported by MSIT, Korea, under [RS-2022-00164860] and [RS-2024-00418641] supervised by IITP.
\clearpage
{
    \small
    \bibliographystyle{ieeenat_fullname}
    \bibliography{main}
}
\clearpage
\clearpage
\setcounter{page}{1}
\maketitlesupplementary

\renewcommand\thesection{\Alph{section}}
\renewcommand{\thefigure}{\Alph{figure}}
\renewcommand{\thetable}{\Alph{table}}
\setcounter{section}{0}
\setcounter{figure}{0}
\setcounter{table}{0}

\section{PerMo Dataset Details}
\subsection{Motion Capture Environment}

\begin{figure}[b]
\vspace{-0.3cm}
\centering
    \includegraphics[width=\linewidth]{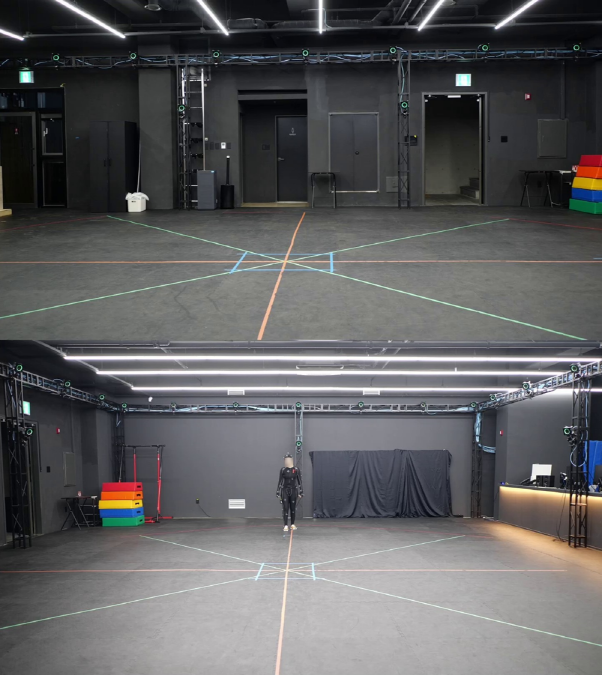}
    \caption{ Side and front views of a motion capture studio equipped with OptiTrack cameras
    }
    \label{fig_sup:studio}
\end{figure}

The dataset was captured in a studio shown in Fig.~\ref{fig_sup:studio} with a floor area of 12m x 10m using 33 OptiTrack\footnote{https://www.optitrack.com} cameras: 25 PrimeX 22 and 8 Prime17W. The PrimeX 22 cameras record at a resolution of $2048 \times 1088$ with a $79 ^{\circ}$ field of view, while the Prime17W cameras record at a resolution of $1664 \times 1088$ with a $70 ^{\circ}$ field of view. All videos were recorded at 120 fps.

\subsection{Persona and Content Categories}
\begin{table}[]
\caption{Style categories included in the PerMo dataset}
\centering
\resizebox{\linewidth}{!}{%
    \begin{tabular}{l|l}
    \toprule[1pt]
    \textbf{Parent Category} & \textbf{Style Category}                                                                                                          \\ \midrule
    Age            & Childish, Neutral, Old, Teenage                                                                                                     \\ \midrule
    Character      & \begin{tabular}[c]{@{}l@{}}Ballerina, Hulk, Monkey, Ninja,  \\ Penguin, Robot, SWAT, Waiter, Zombie\end{tabular}           \\ \midrule
    Condition      & \begin{tabular}[c]{@{}l@{}}Arm-aching, Drunken, Exhausted, \\ Head-aching, Healthy, Leg-aching,\\ Text-necked\end{tabular} \\ \midrule
    Emotion        & \begin{tabular}[c]{@{}l@{}}Angry, Fearful, Happy, Sad, \\ Strained, Surprising\end{tabular}                                \\ \midrule
    Traits    & Elegant, Shy, Silly, Uppity                                                                                                \\ \midrule
    Surroundings   & \begin{tabular}[c]{@{}l@{}}Cold, Crowded, Muddy-floor, \\ Unpleasant-floor\end{tabular}                                    \\ \bottomrule[1pt]
    \end{tabular}
}

\label{tab_sup:persona}
\end{table}
\begin{table}[]
\caption{Content categories included in the PerMo dataset}
\centering
\resizebox{0.75\linewidth}{!}{%
    \begin{tabular}{l|l}
    \toprule[1pt]
    \textbf{Parent Category} & \textbf{Content Category}                                                                                                          \\ \midrule
    Leg Action      & Kick, Kick Something \\\midrule
    Arm Action      & Punch, Throw, Wave \\\midrule
    Ground Locomotion        & Walk, Run, Transition \\\midrule
    Leaping Locomotion   & Jump, Hop\\ 
    \bottomrule[1pt]
    \end{tabular}
}

\label{tab_sup:content}
\end{table}

PerMo was captured by five professional actors, including two women (Actors 1 and 2) and three men (Actors 3, 4, and 5), each with diverse body shapes and heights. The 34 style categories we captured are organized into six groups, as shown in Table~\ref{tab_sup:persona}. Since each of the five actors performed all 34 styles, a total of 170 personas were created. Additionally, for each persona, we recorded motion data for 10 diverse contents to evenly represent full-body movement. These 10 contents fall into one of four action types: \textit{leg actions}, \textit{arm actions}, \textit{ground locomotion}, and \textit{leaping locomotion}, as illustrated in Table~\ref{tab_sup:content}.
Note that actions that were not effective in representing the characteristics of each style were excluded. For instance, in the `muddy-floor' style, the `throw' action was omitted because it does not effectively convey the texture of a sticky floor.

\begin{figure*}
\centering
    \includegraphics[width=0.85\linewidth]{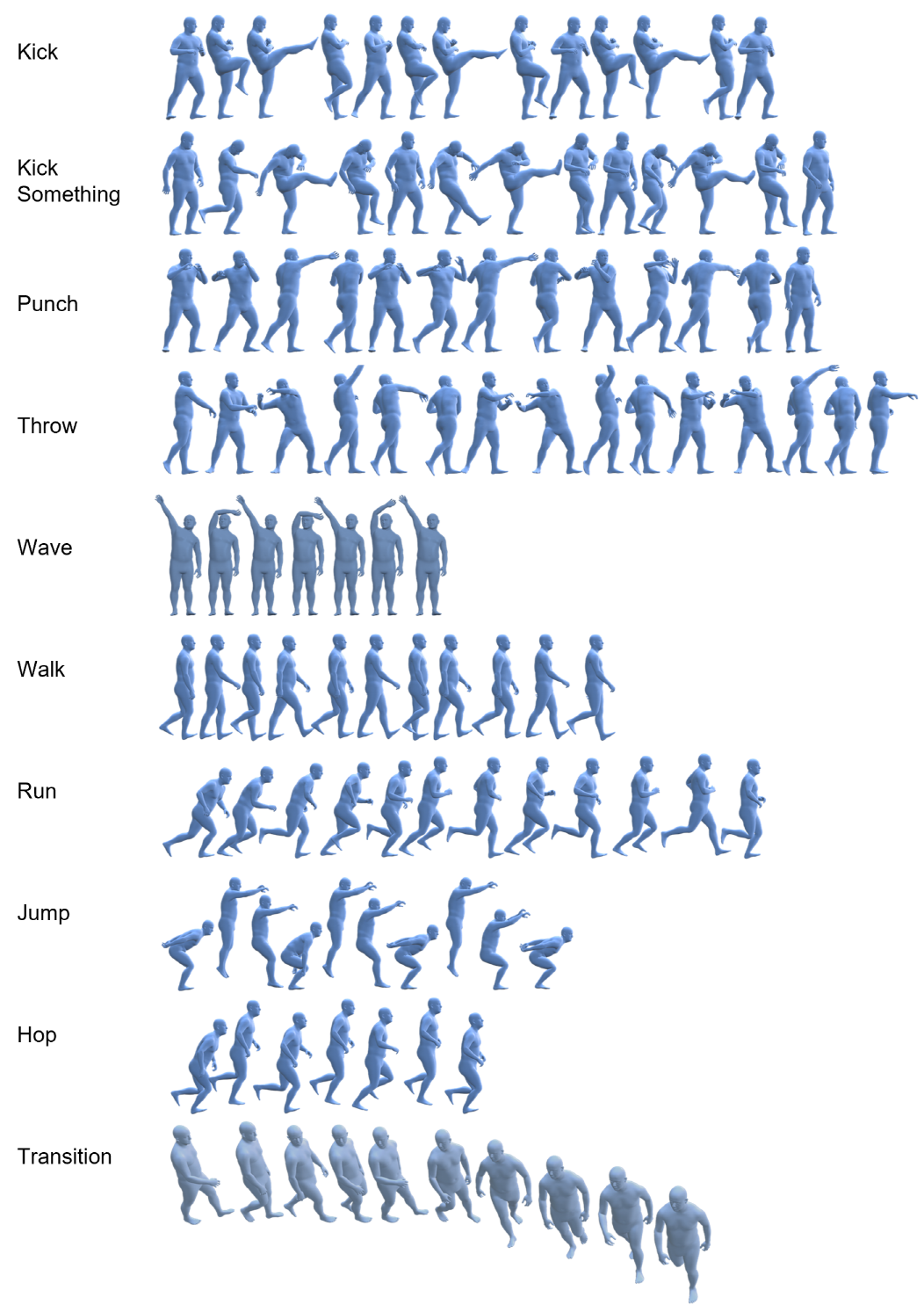}
    \caption{Examples of content types in the PerMo dataset. Please refer to the attached video for a more detailed visualization}
    \label{fig_sup:contents}
\end{figure*}

\begin{figure*}
\centering
    \includegraphics[width=0.9\linewidth]{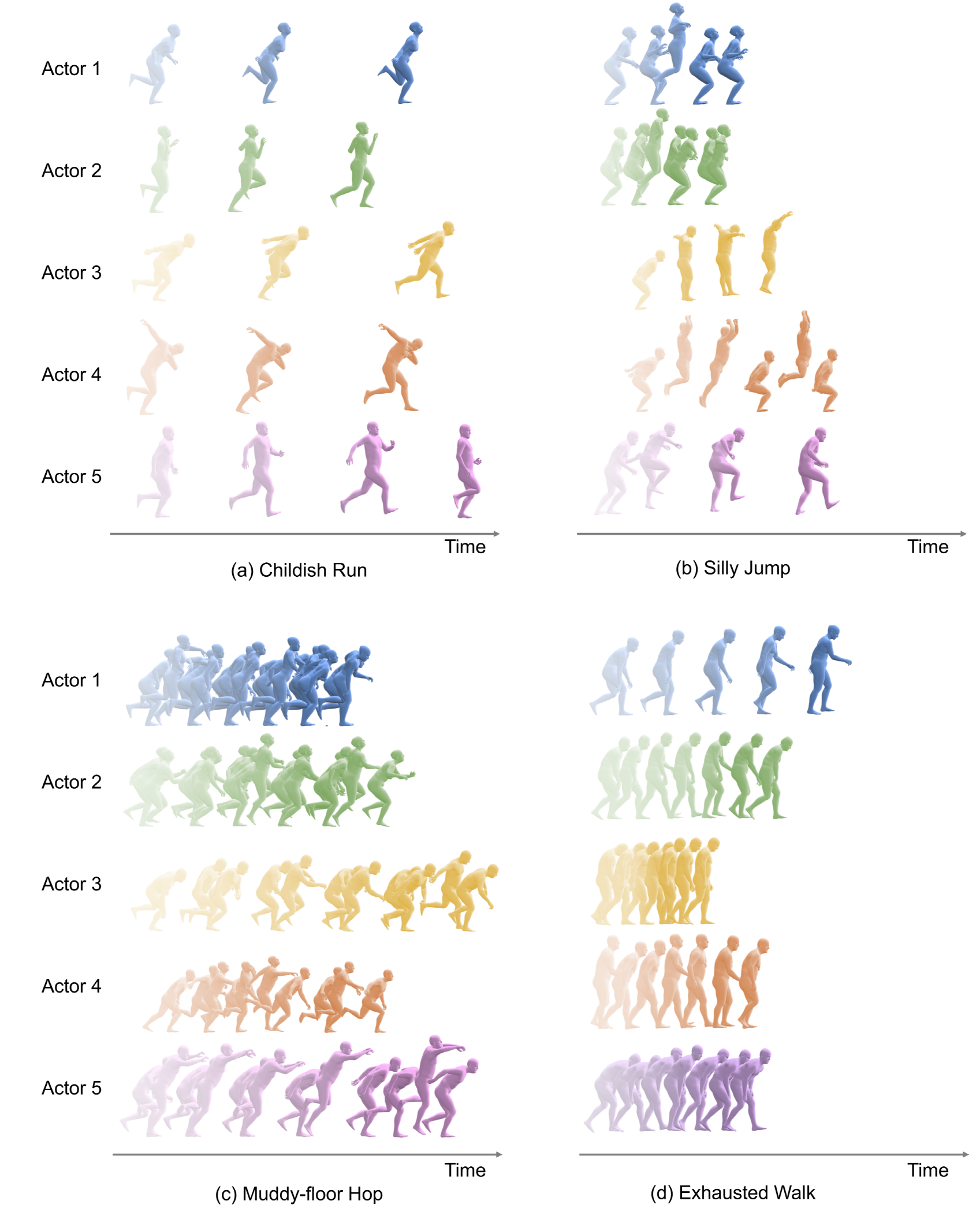}
    \caption{Examples of personas in the PerMo dataset. Even for the same style and content, each actor portrays a different persona. Please refer to the attached video for a more detailed visualization}
    \label{fig_sup:persona}
\end{figure*}

Examples of the content motions are shown in Fig.\ref{fig_sup:contents}, and examples of each actor's personas are presented in Fig.\ref{fig_sup:persona}. Additional examples can be viewed in the attached video.

\subsection{Motion Capture Instructions}
We provided instructions to ensure that each actor could effectively embody their unique persona.
First, we presented the style categories and gave each actor time to consider how they would express them.
Actors were given as much freedom as possible to express their personas.
Atomic motions (contents) within the same persona were performed consecutively to ensure a consistent persona was conveyed across those motions. When selecting the 34 styles, we took the actor's opinion into account and avoided choosing styles that could be similarly expressed in the motion, opting instead for distinct ones.

However, for the content, stricter guidelines were provided to ensure all actors performed actions within a common framework, allowing the differences between personas to be more clearly highlighted.
Among the content types, \textit{leg action} and \textit{arm action} are stationary motions performed while standing, whereas \textit{ground locomotion} and \textit{leaping locomotion} involve movements across a wide range.
For stationary motions, we marked the exact center of the studio and ensured the motions were performed at that location.
For motions requiring movement, we placed a marker 4 meters away from the center. Actors started their motions from the edge of one side of the studio, and during post-processing, the sequence was cropped to begin when the actor stepped on the marker.
Each motion was recorded in four to five takes for the same persona and content type.

We predefined the repetition count for each content type and cropped the data during post-processing to include only the specified number of repetitions.
\textit{Leg action} and \textit{arm action} were repeated three times each, while \textit{ground locomotion} and \textit{leaping locomotion} were repeated five times each.
All motions were recorded with a focus on the right side. For example, \textit{leg action} and \textit{arm action} were performed using the right arm and right leg. \textit{Ground locomotion} started with stepping on the marker with the right foot. The `Transition' action involved turning toward the right, and the `Hop' action was performed using the right leg. For training, the data was augmented by flipping left and right, resulting in a total of 13,220 motion samples.

\begin{figure}
    \centering
    \includegraphics[width=\linewidth]{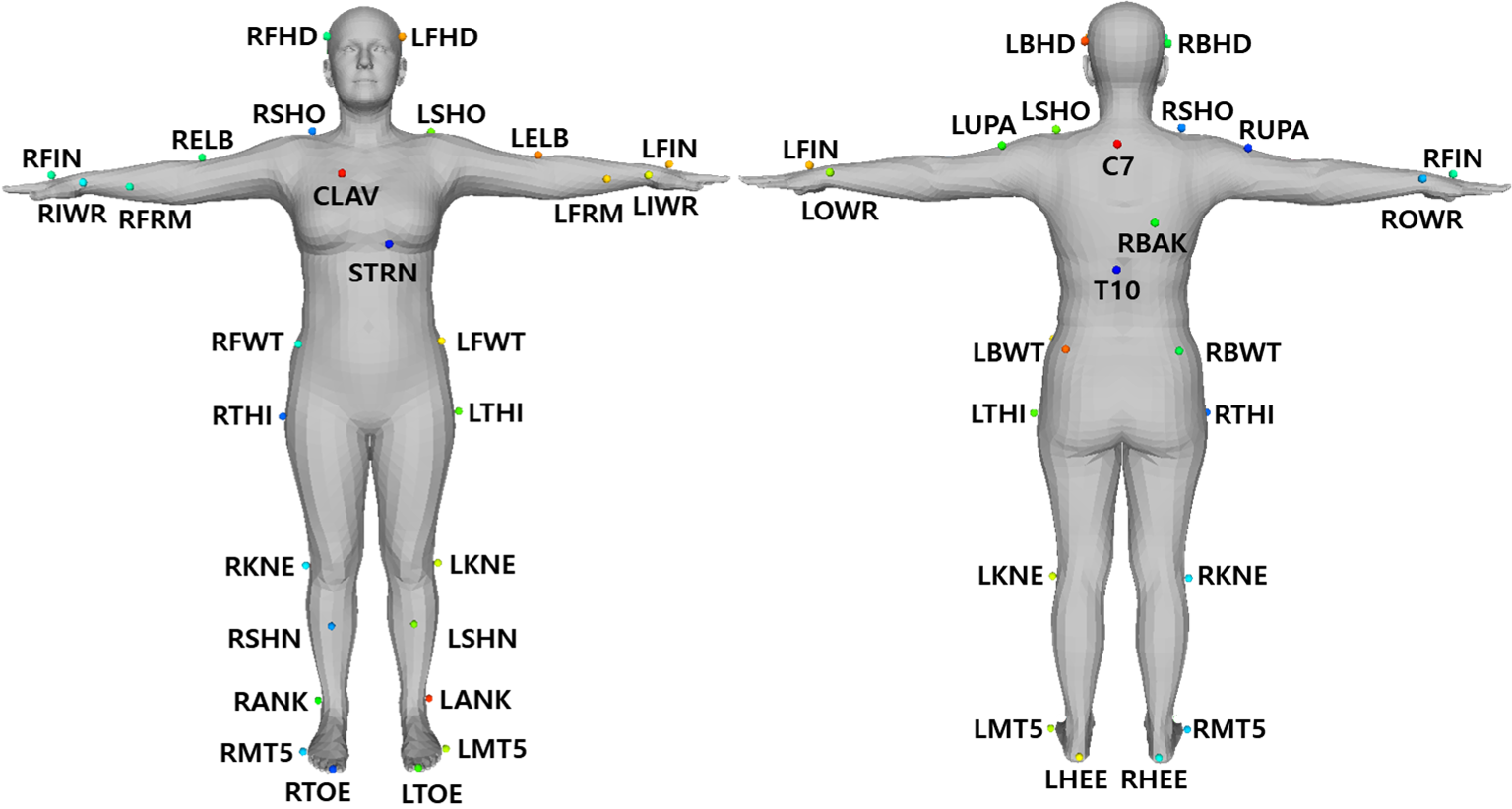}
    \caption{Positions of the 41 markers}
    \label{fig_sup:marker}
\end{figure}

\begin{figure}
    \centering
    \includegraphics[width=\linewidth]{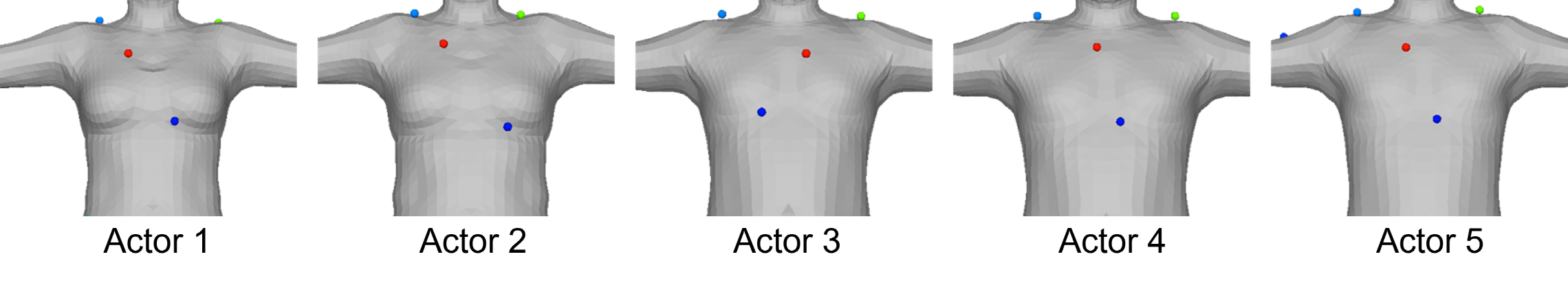}
    \caption{Positions of the CLAV and STRN markers of each actor. The red dot on the upper chest represents CLAV, and the blue dot on the lower chest represents STRN}
    \label{fig_sup:marker_diff}
    \vspace{-0.3cm}
\end{figure}


\begin{figure}
    \centerline{\includegraphics[width=8.5cm, height=5cm]{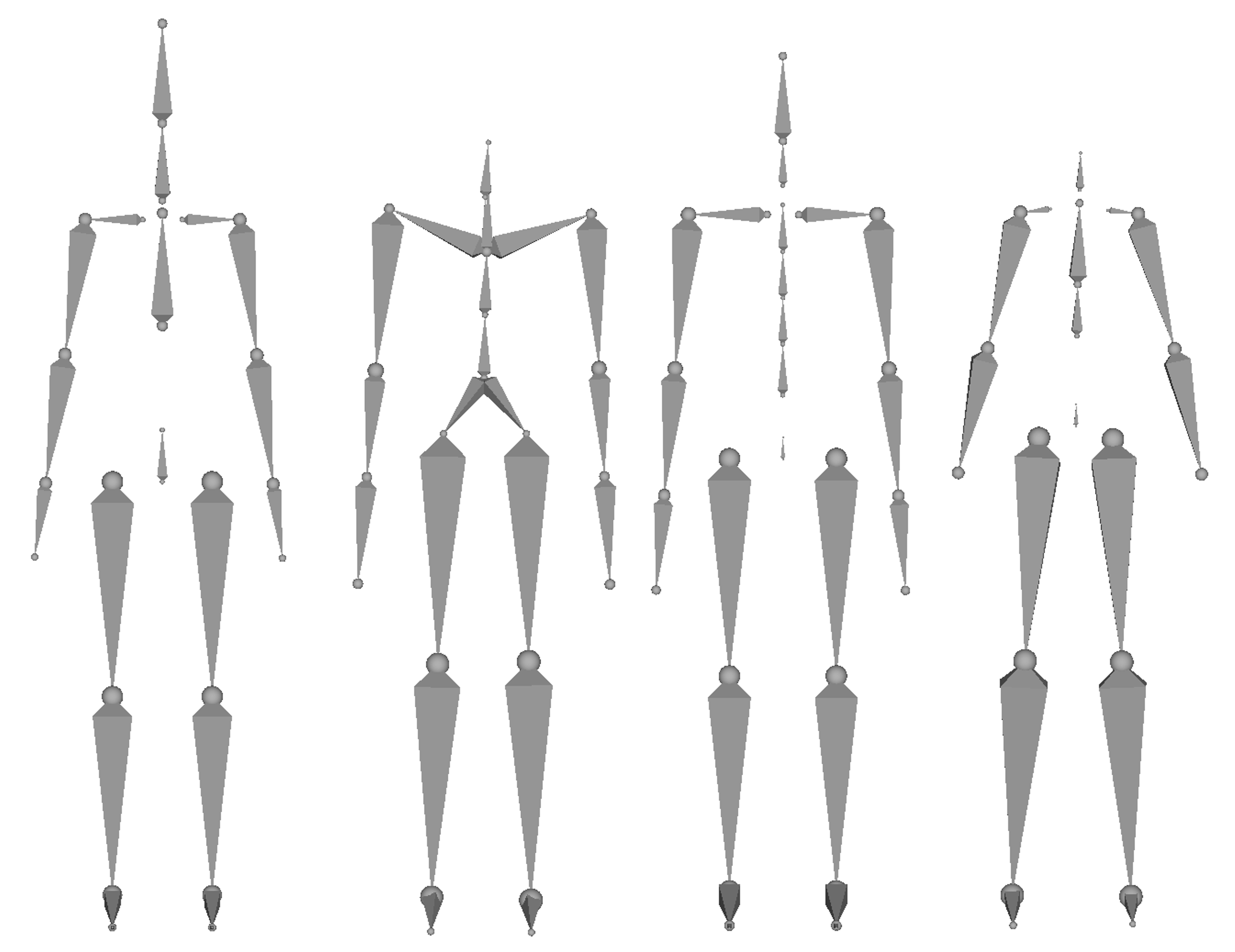}}
    \begin{subfigure}[c]{0.26\linewidth} 
        \caption{}
    \end{subfigure}
    \begin{subfigure}[c]{0.24\linewidth} 
        \caption{}
    \end{subfigure}
    \begin{subfigure}[c]{0.23\linewidth} 
        \caption{}
    \end{subfigure}
    \begin{subfigure}[c]{0.23\linewidth} 
        \caption{}
    \end{subfigure}
    \caption{The skeleton structure of each dataset. (a) PerMo (b) Xia~\cite{xia2015realtime} and BFA~\cite{aberman2020unpaired} (c) 100 STYLE~\cite{mason2022real} (d) BN~\cite{kobayashi2023motion} }
    \label{fig_sup:bone}
\end{figure}
\begin{figure}
    \centering
    \includegraphics[width=\linewidth]{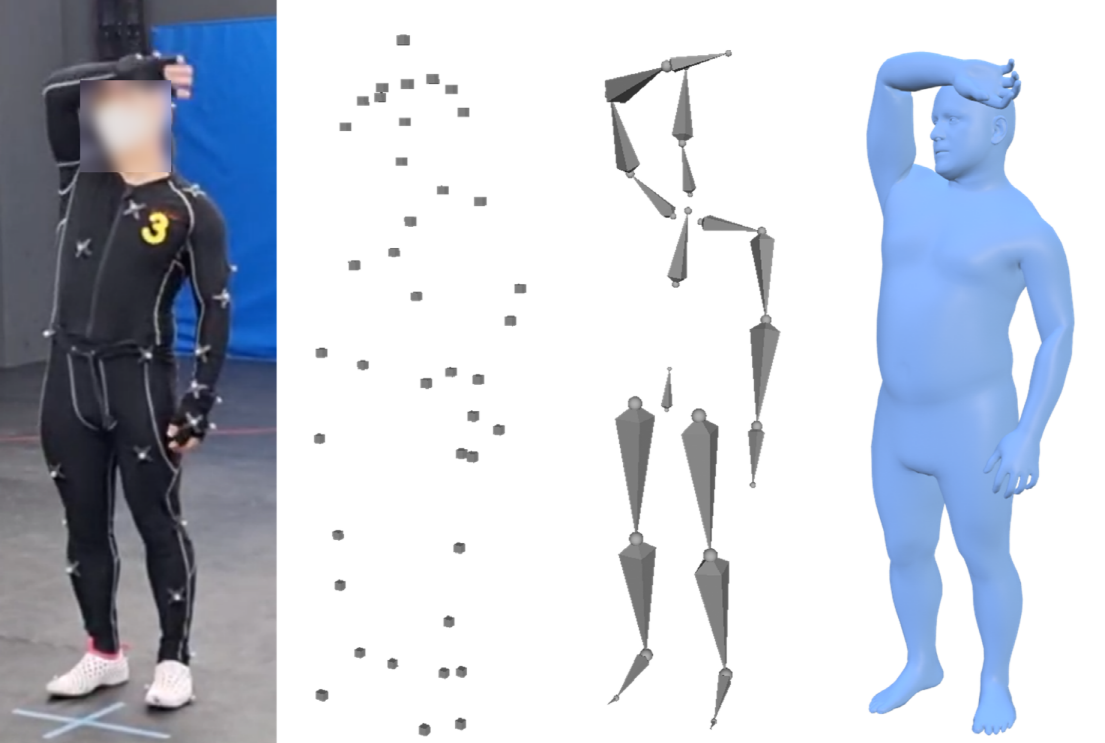}
    \begin{subfigure}[c]{0.2\linewidth} 
        \caption{}
    \end{subfigure}
    \begin{subfigure}[c]{0.2\linewidth} 
        \caption{}
    \end{subfigure}
    \begin{subfigure}[c]{0.2\linewidth} 
        \caption{}
    \end{subfigure}
    \begin{subfigure}[c]{0.2\linewidth} 
        \caption{}
    \end{subfigure}
    \caption{Example of the data formats. (a) RGB image (b) Marker (c) Skeleton (d) Mesh}
    \label{fig_sup:data_format}
\end{figure}


\subsection{Data Format and Post-Processing}

Each actor was equipped with 41 optical markers during the recording process. The positions of these markers are shown in Figure \ref{fig_sup:marker}. To distinguish between actors during data refinement, the positions of the CLAV and STRN markers vary slightly depending on the actor, as illustrated in Figure \ref{fig_sup:marker_diff}. The marker sequences are stored in C3D files.

\begin{table*}[]
\caption{Description of data format and naming conventions }
\centering
\resizebox{0.77\linewidth}{!}{%
    \begin{tabular}{l|l|p{7cm}}
    \toprule[1pt]
    \textbf{Data Type} &  \textbf{File Name} & \textbf{Description} \\ \midrule[1pt]
    Marker & [style]\_[content]\_[actor]\_[take].c3d & 41 markers for each frame \\ \midrule
    Skeleton & [style]\_[content]\_[actor]\_[take].bvh & 20 bones for each frame \\\midrule
    Mesh & [style]\_[content]\_[actor]\_[take].npz & SMPL-H pose data \\\midrule
    Mesh Shape & shape\_[style]\_[content]\_[actor].npz & SMPL-H shape data for each processing group \\\midrule
    Rendered Mesh & [style]\_[content]\_[actor]\_001.mp4 & Rendered video of the mesh data\\ \bottomrule[1pt]
    \end{tabular}
}

\label{tab_sup:format}
\end{table*}

From the marker data, skeleton information comprising 20 bones is generated and saved in BVH files using OptiTrack software. The skeleton structure of PerMo is depicted in Figure~\ref{fig_sup:bone} (a). 
Figures~\ref{fig_sup:bone} (b), (c), and (d) show the skeleton structures of other motion style datasets. As shown, the skeleton structures vary across datasets, making it challenging to use them together for training.
To address this, we convert the data into the standardized SMPL, a widely used 3D human mesh format. This unified format allows compatibility with other large-scale motion datasets such as AMASS~\cite{mahmood2019amass} and HumanML3D~\cite{guo2022generating}. 

We use MoSh++~\cite{mahmood2019amass} to convert optical marker data into SMPL-H format. In the first step of MoSh++, the shape parameters are estimated. To achieve this, we grouped motions that share the same persona and content (4 or 5 takes), and then optimized the mesh for each group. 
Since the motion capture sessions were conducted over multiple days, slight variations in marker positions occurred between sessions. However, for motions within the same group, the marker positions remained consistent, enabling effective mesh optimization.
A subset of frames from the motion sequences within each group was extracted, and the shape parameters were estimated from these frames.
Subsequently, using the extracted shape parameters, the poses for each frame of the motion sequences were optimized.

\begin{figure}[t]
\centering
    \includegraphics[width=0.9\linewidth]{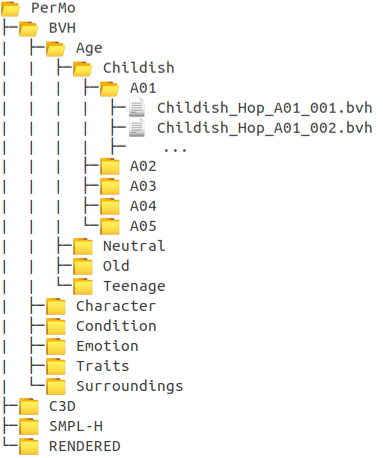}
    \caption{Folder structure of the PerMo dataset
    }
    \label{fig_sup:structure}
\end{figure}

To make it easier for users to check the motions, we provide rendered mesh videos for the first take of each group.
Examples of the described data formats are shown in Fig.~\ref{fig_sup:data_format}, and the summary and naming conventions of the released data are presented in Table~\ref{tab_sup:format}.
In addition, the folder structure of the PerMo dataset is shown in Fig.~\ref{fig_sup:structure}.

\subsection{Data Validation}
We have prepared to release clean data by conducting a rigorous validation of the acquired motion data. The validation process considers the following four factors: (1) the cleanliness of the skeleton data (free from distortions), (2) the accuracy of motion cropping, (3) the presence of missing markers and bones, and (4) the synchronization between marker and skeleton data.

For (1), any twisted skeletons are manually corrected by experts during data validation. Regarding (2), each motion file is manually verified by at least two reviewers, who check whether the actor’s starting position in \textit{ground locomotion} is accurate and whether the correct number of repetitions is cropped. For (3) and (4), custom validation scripts were developed to perform automated checks.

\subsection{Text Description}
\begin{figure*}
\centering
    \includegraphics[width=\linewidth]{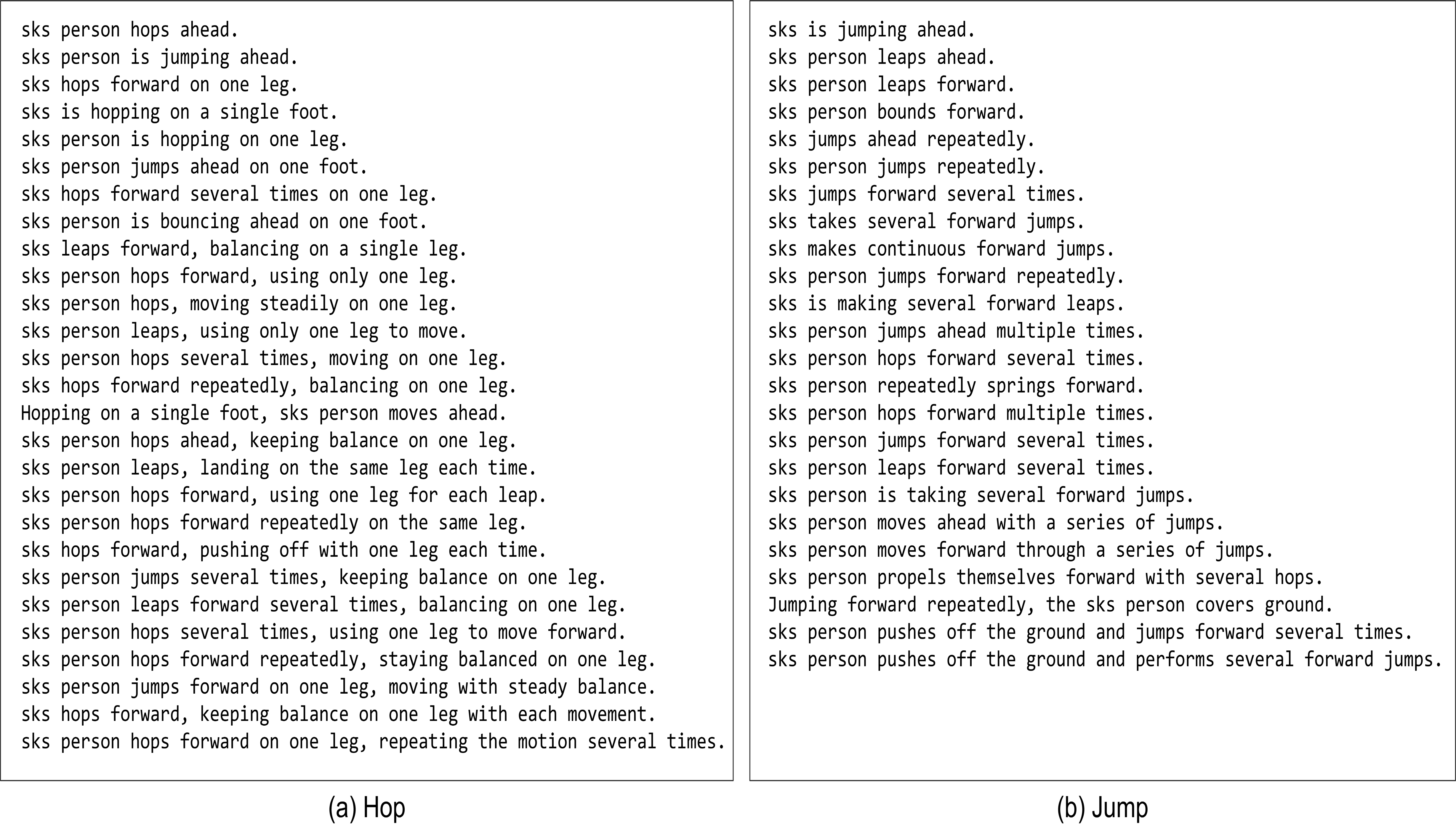}
    \caption{Examples of text descriptions in PerMo dataset}
    \label{fig_sup:text}
\end{figure*}

Examples of text descriptions included in the PerMo dataset are presented in Fig.~\ref{fig_sup:text}.
To generate diverse descriptions, we first provided ChatGPT with detailed explanations of the motions. For example, for the `Hop' motion, we used a description like: \textit{``In the motion, one person hops forward on one leg. The person hops several times."} We then instructed ChatGPT to create 20–30 variations of sentences, ranging from short and simple high-level descriptions to long and detailed low-level ones. As a result, we obtained descriptions of the `Hop' motion at various levels of detail, as shown in Fig.~\ref{fig_sup:text} (a).
Additionally, we represented the placeholder word [P*] within the sentences using \textit{sks}~\cite{ruiz2023dreambooth}. When extracting sentence features, the word embedding corresponding to the \textit{sks} index is replaced with $P^*$.

\begin{figure*}
\centering
    \includegraphics[trim=0 10 0 0, clip, width=\linewidth]{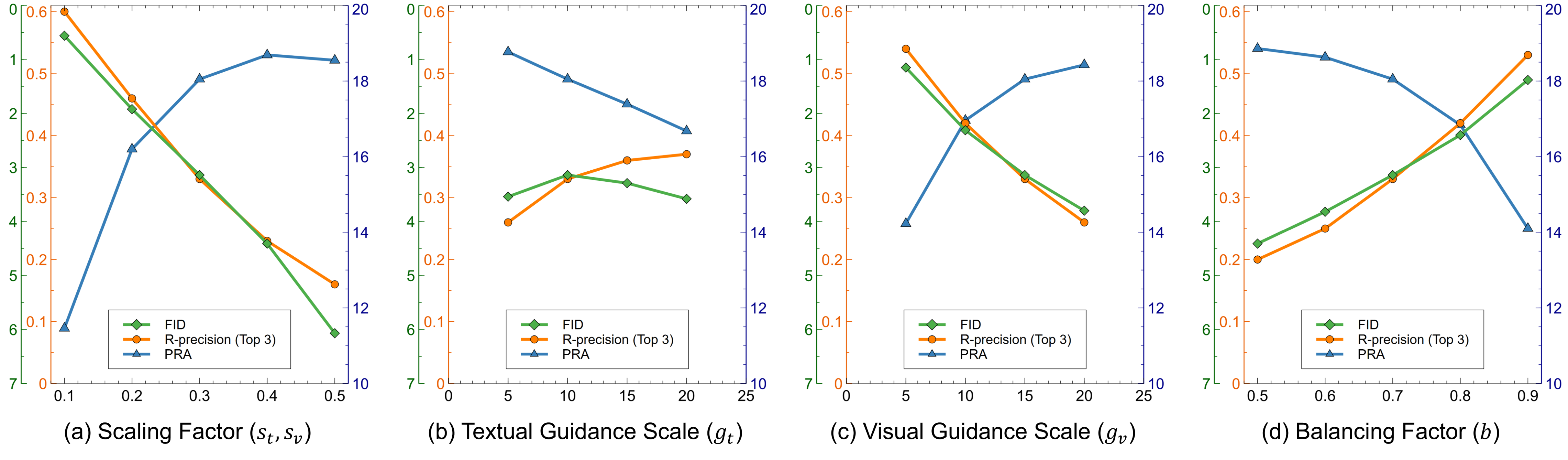}
    \caption{Ablation study for hyperparameters $s_t$, $s_v$, $g_t$, $g_v$, and $b$. Higher positions on the graph indicate better performance across all three metrics, FID, R-precision (Top 3), and PRA.}
    \label{fig_sup:hyperparam}
\end{figure*}

\section{Ablation Study on Hyperparameters}
We conduct an ablation study to examine the effects of hyperparameters $s_t$, $s_v$, $g_t$, $g_v$, and $b$, as shown in Fig.~\ref{fig_sup:hyperparam}. The experiments were performed using the single-input PersonaBooth.
Fig.~\ref{fig_sup:hyperparam} (a) shows how the FID, R-precision (Top 3), and PRA metrics vary with the scaling factors $s_t$ and $s_v$.
As the scaling factor increases, the influence of personality features becomes more pronounced, leading to higher PRA values. However, PRA values reach a saturation point at scaling factors above 0.4. In contrast, FID and R-precision exhibit a trade-off with PRA, showing a decline in performance as the scaling factor increases.

Fig.~\ref{fig_sup:hyperparam} (b) shows the results for the textual guidance scale $g_t$. As $g_t$ increases, the generated motions align more closely with the prompt, leading to improved R-precision. Conversely, PRA decreases, demonstrating a trade-off with R-precision. The FID metric performs best around $g_t = 10$.
Fig.~\ref{fig_sup:hyperparam} (c) presents the results for the visual guidance scale $g_v$. As $g_v$ increases, the generated motions more accurately reflect the features of the input motion, resulting in higher PRA values. Conversely, both FID and R-precision performance decrease.
Fig.~\ref{fig_sup:hyperparam} (d) shows the results for the balancing factor $b$. Larger $b$ values place more emphasis on the text, leading to a decrease in PRA but an improvement in both FID and R-precision.
In selecting hyperparameters, we prioritized reflecting the persona, as the main objective of this task is personalization.
Therefore, we aimed to maintain a high level of PRA while also achieving favorable FID and R-precision scores.

\section{Ablation Study on the Number of Inputs}
\begin{figure}
\centering
    \includegraphics[trim=0 10 0 0, clip, width=0.9\linewidth]{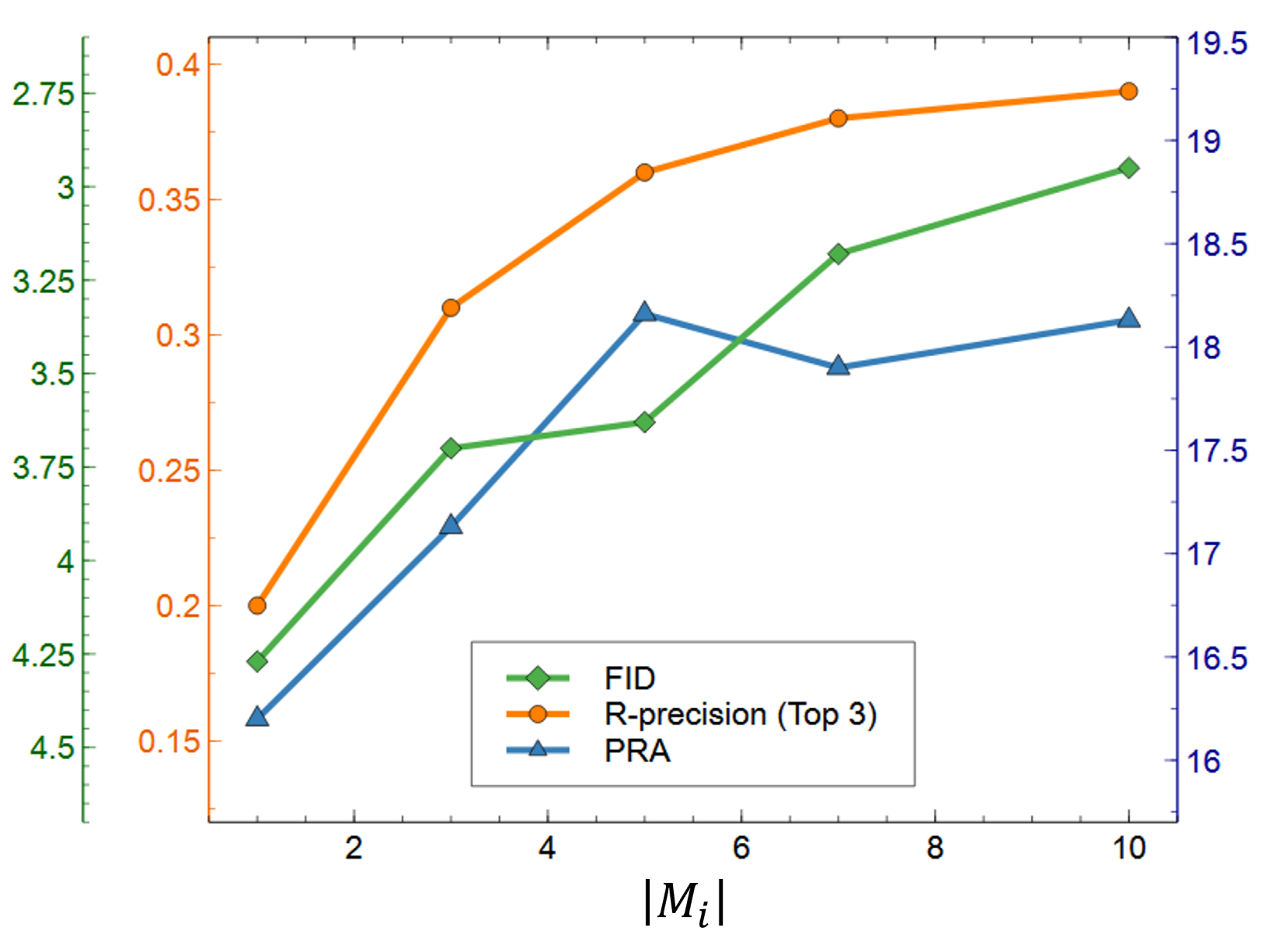}
    \caption{
    Ablation study on the number of input motions. Higher positions on the graph indicate better performance.
    }
    \label{fig_sup:num_input}
\end{figure}

An ablation study on the number of inputs is conducted in the multiple input (MI) setting. As $|M_i|$ increases, all metrics demonstrate an increasing trend. This can be attributed to the higher likelihood of encountering motions that are more contextually aligned with the prompt as the diversity of input motions grows.

\section{Persona Recognition Accuracy Details}
For PerMo, PRA aims to classify 170 personas. To reduce the burden on a single classifier handling too many classes, we used nine separate classifiers, each responsible for a sub-parent category. Table~\ref{tab_sup:eval_group} shows these sub-parent categories for PRA. For instance, the classifier for the "Age" category handles 20 personas, with each style containing five different personas.

Each classifier is a 2-block transformer-based model, pretrained separately for each sub-parent category. The classifiers are trained using ground truth motions in PerMo paired with persona labels. These classifiers evaluate the generated motions, and their classification accuracy determines the PRA. Note that all 170 personas are combined into a single dataset during training.
For 100Style, Style Recognition Accuracy (SRA) is evaluated across 40 style categories, excluding content-oriented ones. Only one classifier is used for 100Style.

\begin{table}[t!]
\caption{Sup-parent categories for measuring the PRA metric}
\centering
\resizebox{\linewidth}{!}{%
    \begin{tabular}{l|l}
    \toprule[1pt]
    \textbf{Sub-Parent Category} & \textbf{Persona Category}                                                                                                          \\ \midrule
    Age            & Childish, Neutral, Old, Teenage                                                                                                     \\ \midrule
    Character 1      & \begin{tabular}[c]{@{}l@{}}Ballerina, Hulk, Monkey, Ninja\end{tabular} \\ \midrule
    Character 2      & \begin{tabular}[c]{@{}l@{}} Penguin, Robot, SWAT, Waiter, Zombie\end{tabular} \\ \midrule
    Condition 1      & \begin{tabular}[c]{@{}l@{}}Arm-aching, Drunken, Exhausted, \\ Head-aching\end{tabular} \\ \midrule
    Condition 2      & \begin{tabular}[c]{@{}l@{}} Healthy, Leg-aching, Text-necked\end{tabular} \\ \midrule
    Emotion 1        & \begin{tabular}[c]{@{}l@{}}Angry, Fearful, Happy \end{tabular} \\ \midrule
    Emotion 2        & \begin{tabular}[c]{@{}l@{}} Sad, Strained, Surprising\end{tabular} \\ \midrule
    Traits    & Elegant, Shy, Silly, Uppity \\ \midrule
    Surroundings   & \begin{tabular}[c]{@{}l@{}}Cold, Crowded, Muddy-floor, \\ Unpleasant-floor\end{tabular}                                    \\ \bottomrule[1pt]
    \end{tabular}
}

\label{tab_sup:eval_group}
\vspace{-0.2cm}
\end{table}

\clearpage


\end{document}